\documentclass[review]{elsarticle}

\usepackage{color}
\usepackage{graphicx}
\usepackage{tabularx}
\usepackage{multirow}
\usepackage{epstopdf}
\usepackage{amsmath}
\usepackage{amssymb}
\usepackage{bbding}
\usepackage{textcomp}
\usepackage{setspace}
\usepackage{color}
\usepackage{tikz}
\usepackage{subcaption}
\usepackage{pgfplots}
\usepackage{url}

\newcolumntype{?}{!{\vrule width 1pt}}

\journal{Computerized Medical Imaging and Graphics}

\begin{document}

\begin{frontmatter}

\title{Healthy versus pathological learning transferability in shoulder muscle MRI segmentation using deep convolutional encoder-decoders}

\author[l1,l2]{Pierre-Henri Conze \corref{cor1}}
\author[l2,l3,l4]{Sylvain Brochard}
\author[l1,l2]{Val\'erie Burdin}
\author[l5]{Frances T. Sheehan}
\author[l2,l3,l4]{Christelle Pons}

\address[l1]{IMT Atlantique, LaTIM UMR 1101, UBL, Technop\^ole Brest-Iroise, 29238 Brest, France}

\address[l2]{Inserm, LaTIM UMR 1101, IBRBS, 22 rue Camille Desmoulins, 29238 Brest, France}

\address[l3]{Rehabilitation Medicine, University Hospital of Brest, 2 avenue Foch, 29200 Brest, France }

\address[l4]{SSR pediatric, Fondation ILDYS, Ty Yann, rue Alain Colas, 29218 Brest, France}

\address[l5]{Rehabilitation Medicine, NIH, 10 Center Drive, MD 20892, Bethesda, USA \vspace{-0.5cm}}

\cortext[cor1]{corresponding author: \url{pierre-henri.conze@imt-atlantique.fr}}

\begin{abstract}
Fully-automated segmentation of pathological shoulder muscles in patients with musculo-skeletal diseases is a challenging task due to the huge variability in muscle shape, size, location, texture and injury. A reliable automatic segmentation method from magnetic resonance images could greatly help clinicians to diagnose pathologies, plan therapeutic interventions and predict interventional outcomes while eliminating time consuming manual segmentation. The purpose of this work is three-fold. First, we investigate the feasibility of automatic pathological shoulder muscle segmentation using deep learning techniques, given a very limited amount of available annotated pediatric data. Second, we address the learning transferability from healthy to pathological data by comparing different learning schemes in terms of model generalizability. Third, extended versions of deep convolutional encoder-decoder architectures using encoders pre-trained on non-medical data are proposed to improve the segmentation accuracy. Methodological aspects are evaluated in a leave-one-out fashion on a dataset of 24 shoulder examinations from patients with unilateral obstetrical brachial plexus palsy and focus on 4 rotator cuff muscles (deltoid, infraspinatus, supraspinatus and subscapularis). The most accurate segmentation model is partially pre-trained on the large-scale ImageNet dataset and jointly exploits inter-patient healthy and pathological annotated data. Its performance reaches Dice scores of 82.4\%, 82.0\%, 71.0\% and 82.8\% for deltoid, infraspinatus, supraspinatus and subscapularis muscles. Absolute surface estimation errors are all below 83mm$^2$ except for supraspinatus with 134.6mm$^2$. The contributions of our work offer new avenues for inferring force from muscle volume in the context of musculo-skeletal disorder management.

\end{abstract}


\end{frontmatter}

\section{Introduction}
\label{sec:sec1}

The rapid development of non-invasive imaging technologies over the last decades has opened new horizons in studying both healthy and pathological anatomy. As part of this, pixel-wise segmentation has become a crucial task in medical image analysis with numerous applications such as computer-assisted diagnosis, surgery planning, visual augmentation, image-guided interventions and extraction of quantitative indices from images. However, the analysis of complex magnetic resonance (MR) imaging datasets \textcolor{black}{is cumbersome} and time-consuming for radiologists, clinicians and researchers. Thus, computerized assistance methods, including robust automatic image segmentation techniques, are needed to guide and improve image interpretation and clinical decision making.  

\begin{figure}
\vspace{-0.1cm}
\centering \begin{tabular}{cccc}
& \hspace{-0.3cm} \footnotesize axial & \hspace{-0.45cm} \footnotesize coronal & \hspace{-0.45cm} \footnotesize sagittal \vspace{0.05cm} \cr
\rotatebox{90}{\footnotesize \textcolor{white}{----} deltoid} &
\hspace{-0.35cm} \includegraphics[height=1.9cm]{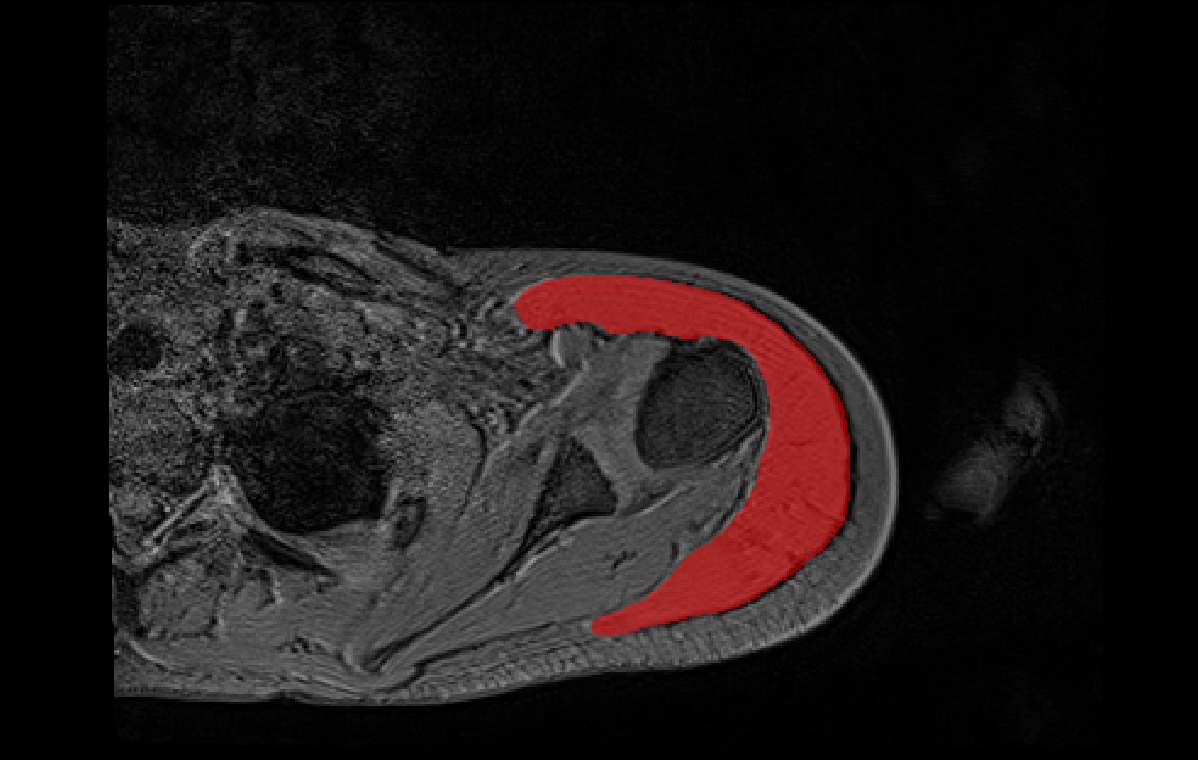} &
\hspace{-0.4cm} \includegraphics[height=1.9cm]{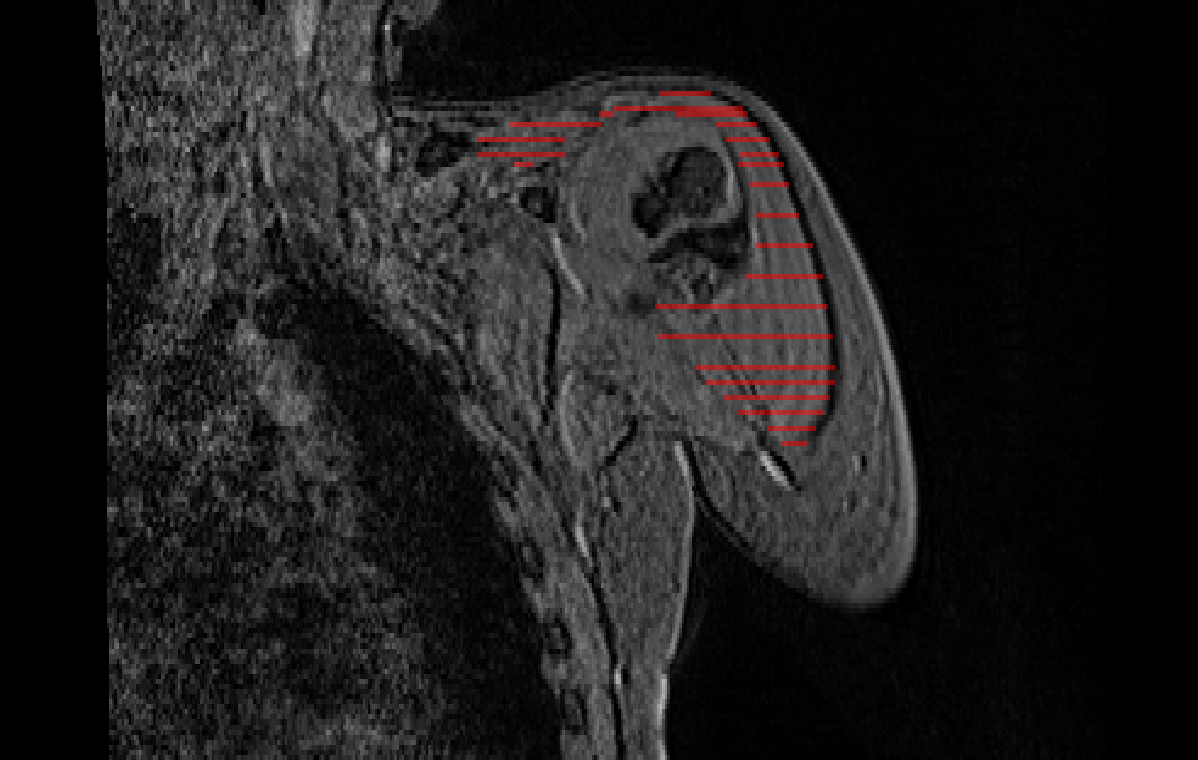} &
\hspace{-0.4cm} \includegraphics[height=1.9cm]{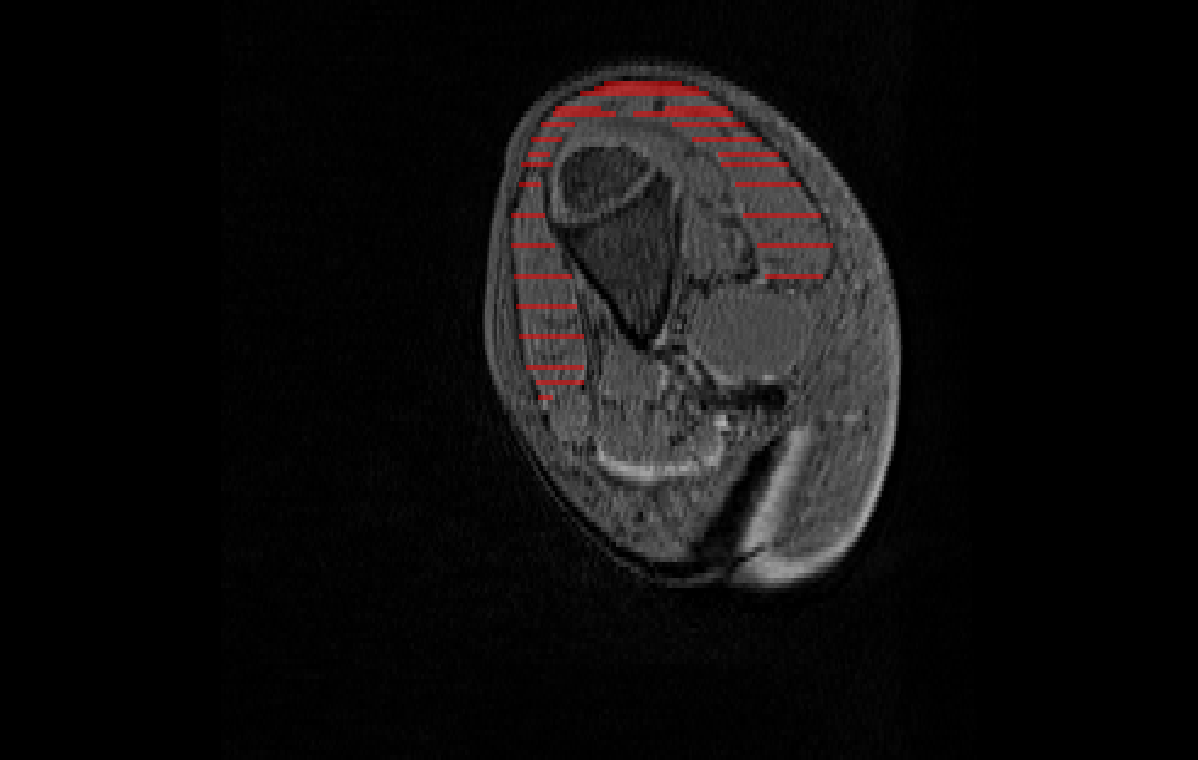} \cr
\rotatebox{90}{\tiny\textcolor{white}{..}\footnotesize infraspinatus} &
\hspace{-0.35cm} \includegraphics[height=1.9cm]{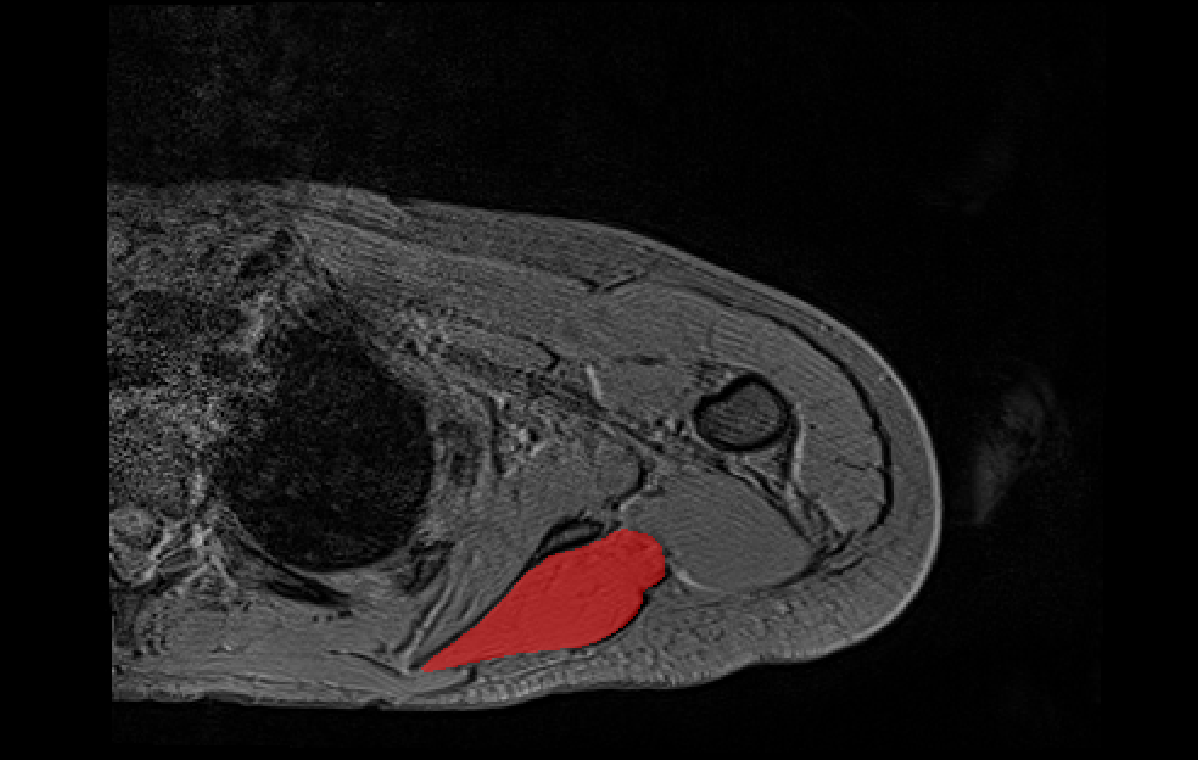} &
\hspace{-0.4cm} \includegraphics[height=1.9cm]{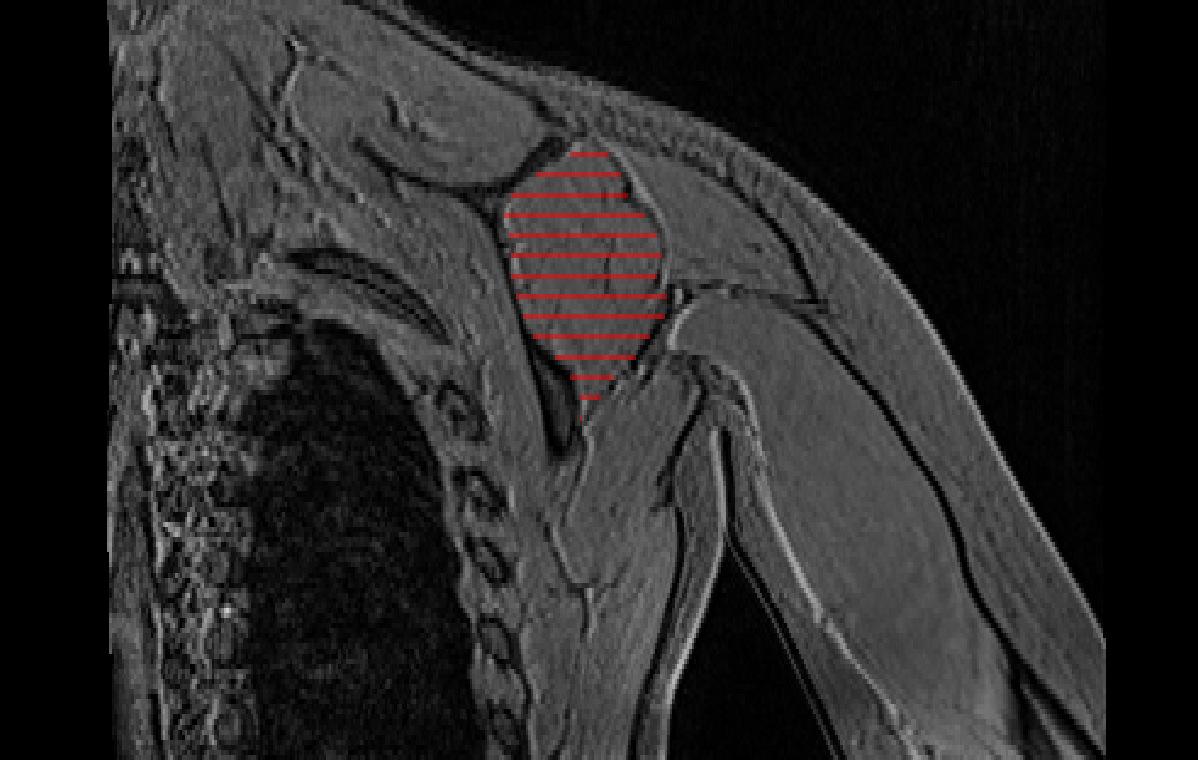} &
\hspace{-0.4cm} \includegraphics[height=1.9cm]{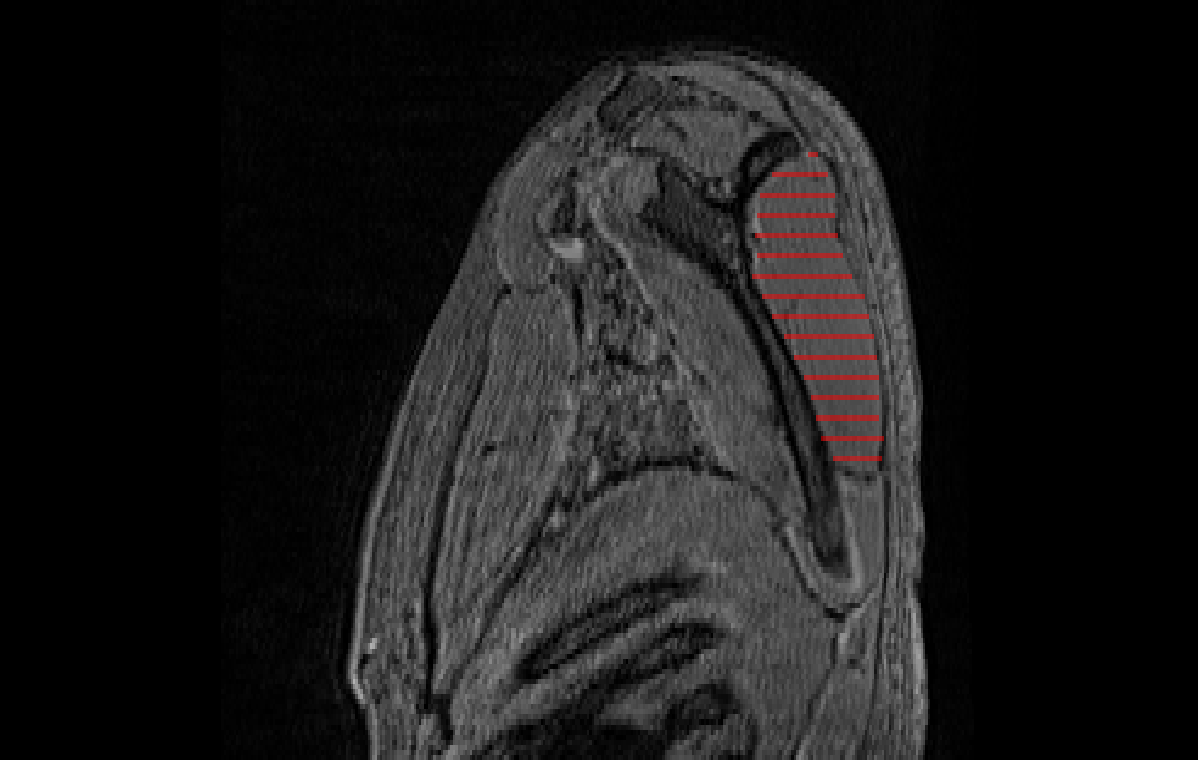} \cr
\rotatebox{90}{\tiny\textcolor{white}{.}\footnotesize  supraspinatus} &
\hspace{-0.35cm} \includegraphics[height=1.9cm]{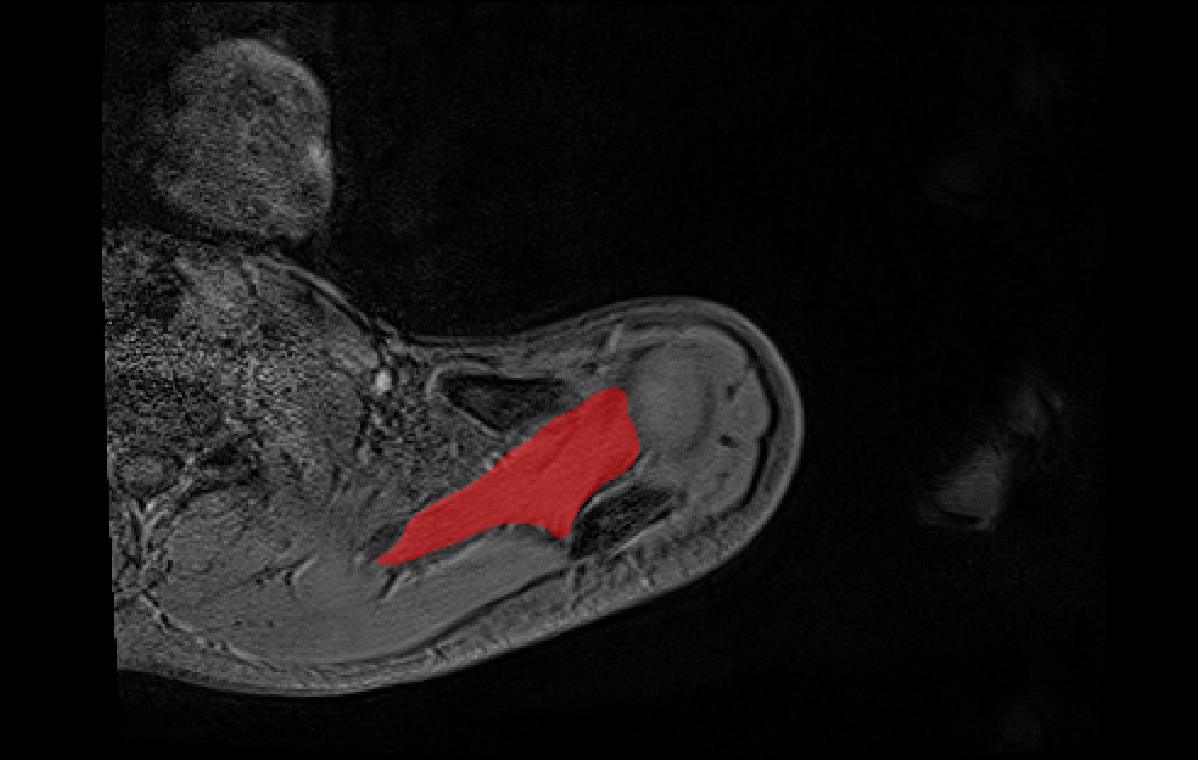} &
\hspace{-0.4cm} \includegraphics[height=1.9cm]{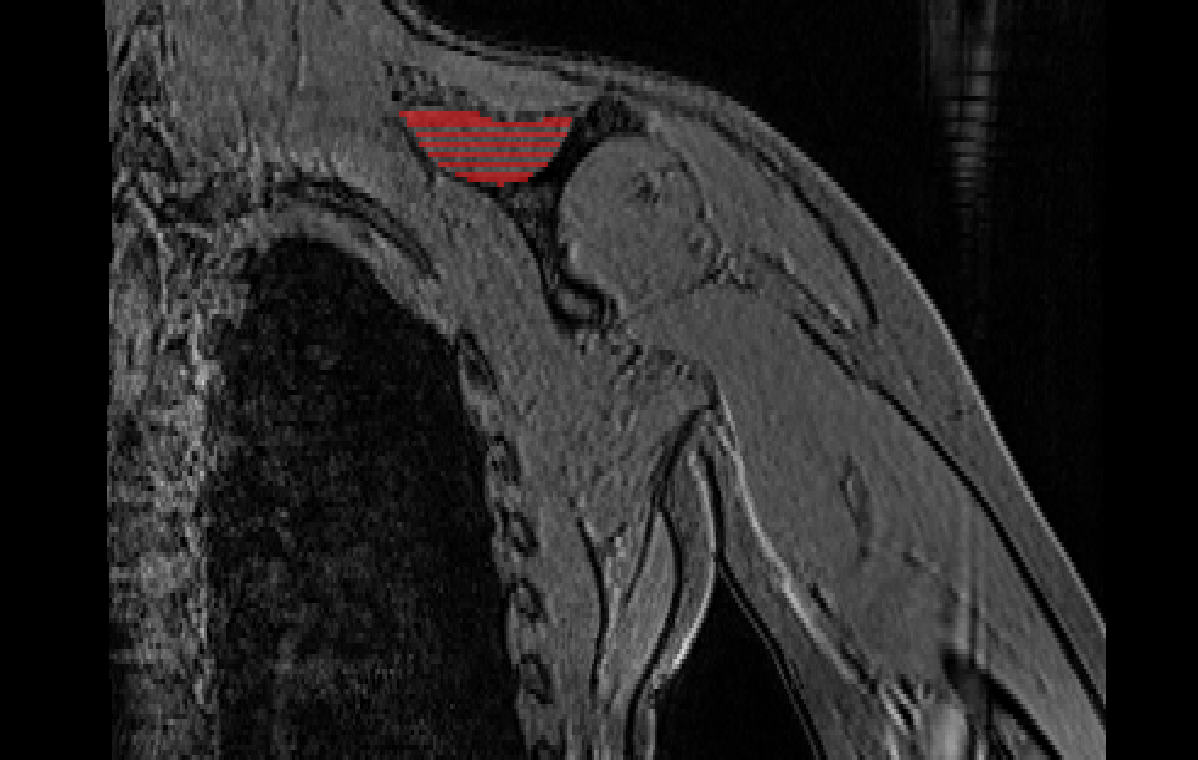} &
\hspace{-0.4cm} \includegraphics[height=1.9cm]{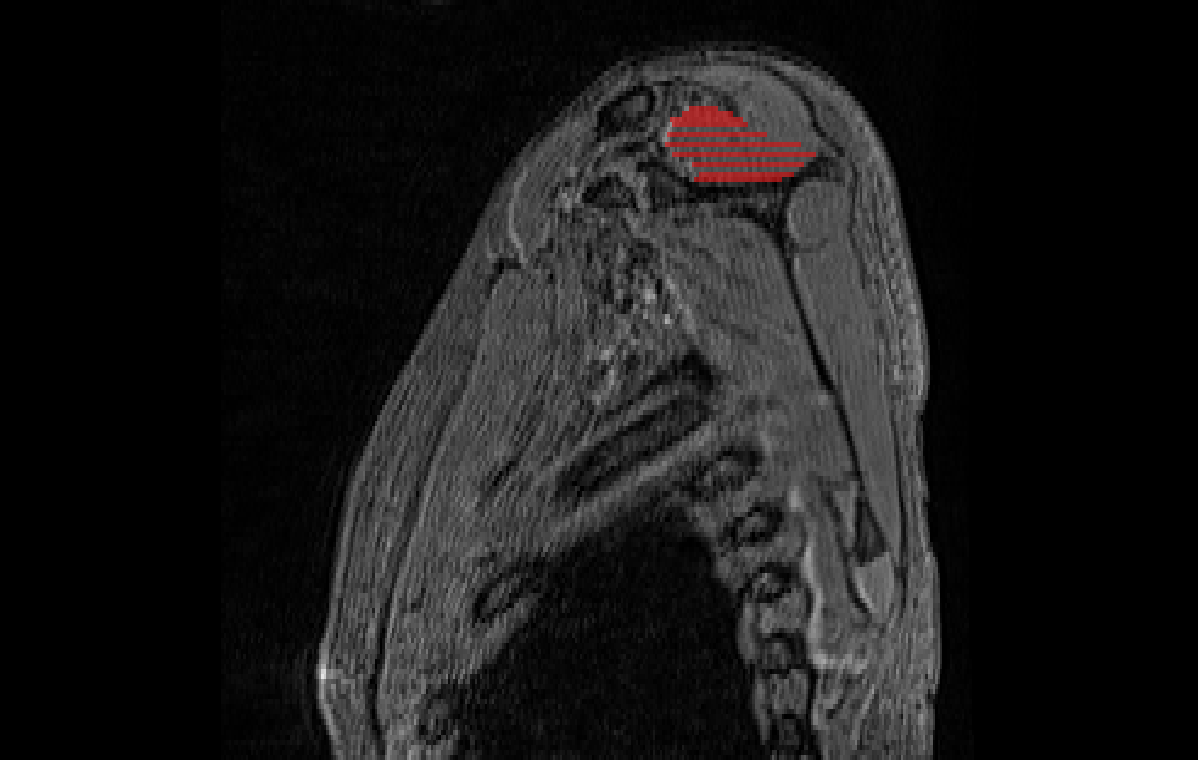} \cr
\rotatebox{90}{\tiny\textcolor{white}{.}\footnotesize subscapularis} &
\hspace{-0.35cm} \includegraphics[height=1.9cm]{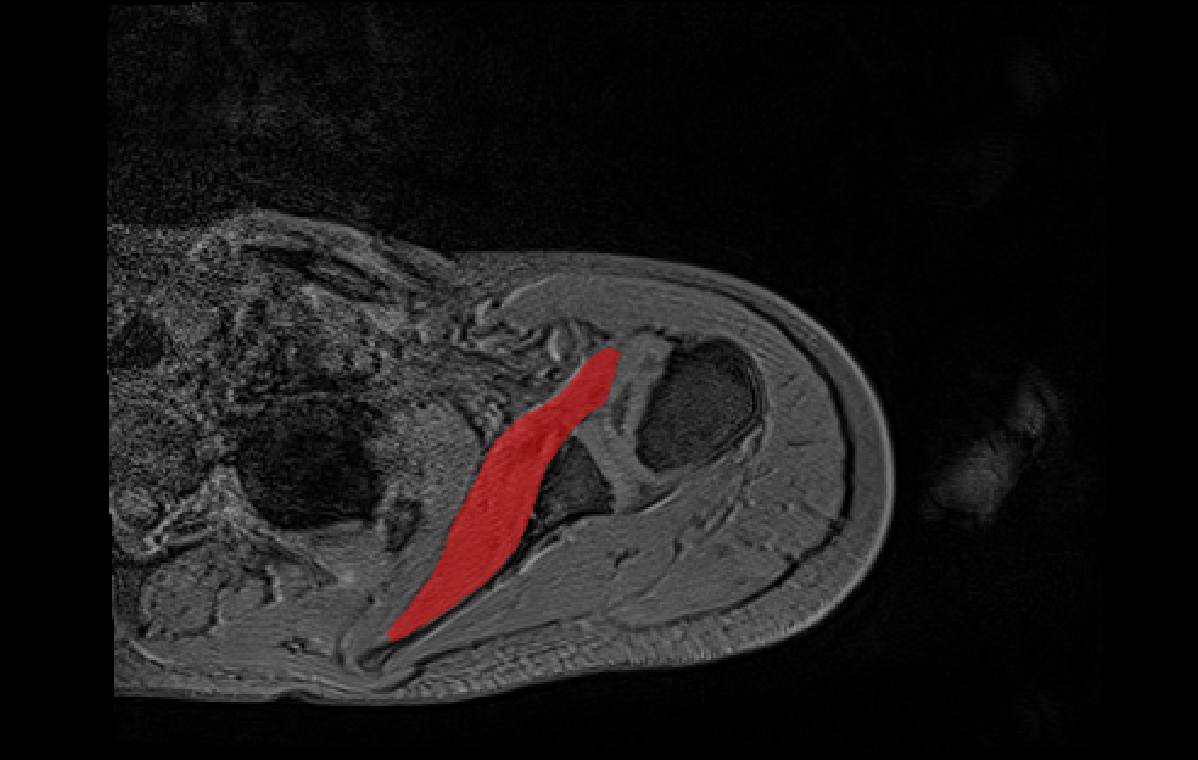} &
\hspace{-0.4cm} \includegraphics[height=1.9cm]{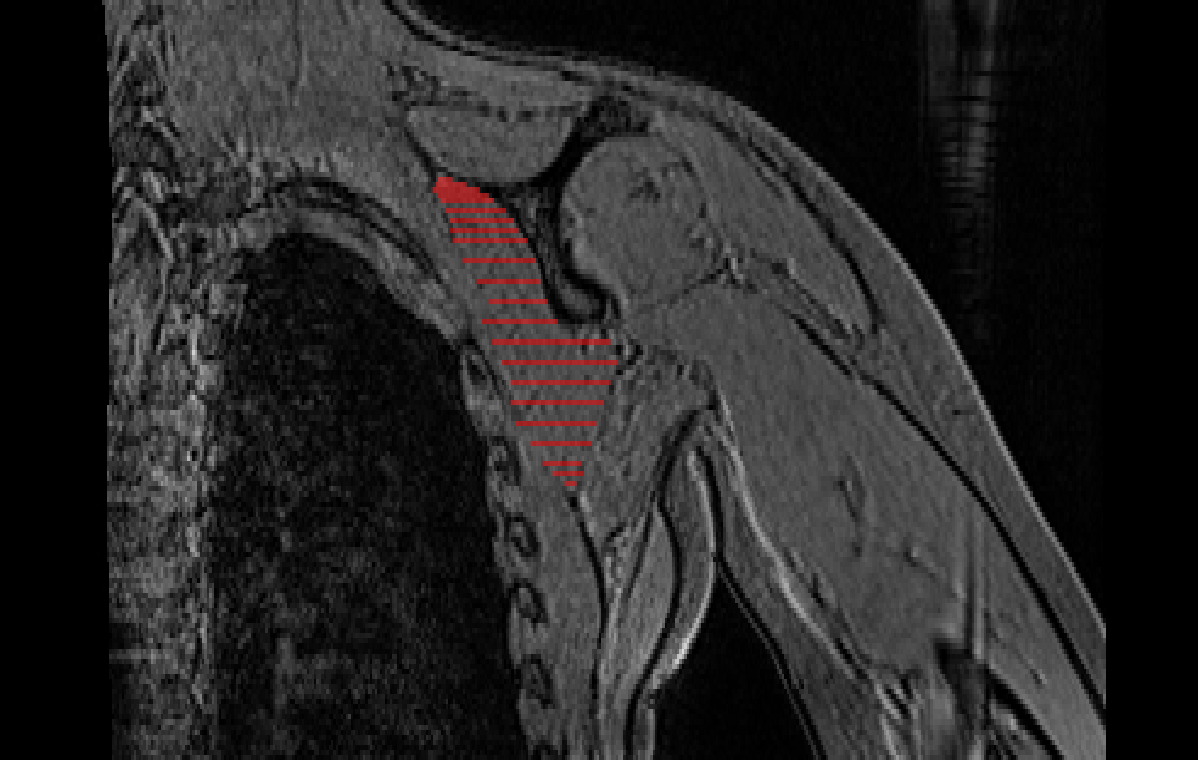} &
\hspace{-0.4cm} \includegraphics[height=1.9cm]{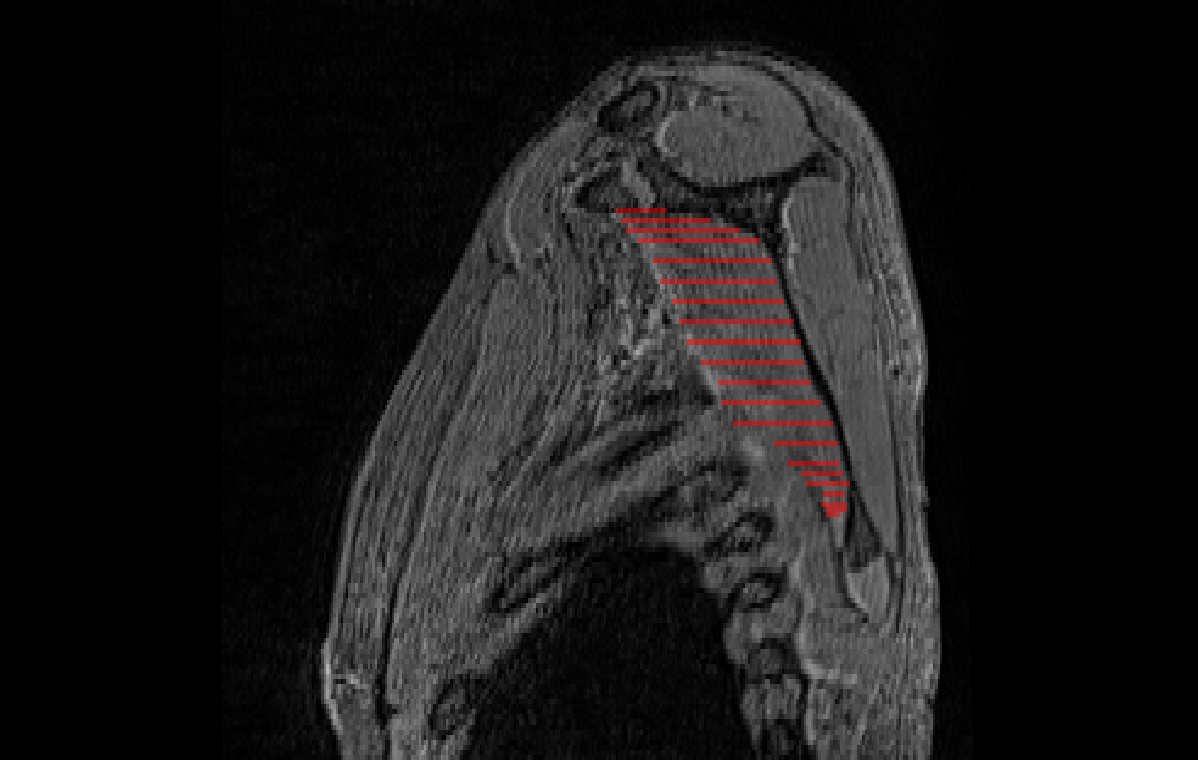} \cr
\end{tabular}
\caption{Groundtruth \textcolor{black}{segmentation} of pathological shoulder muscles including deltoid as well as infraspinatus, supraspinatus and subscapularis from the rotator cuff. Axial, coronal and sagittal slices are extracted from a 3D MR examination acquired for a child with obstetrical brachial plexus palsy.}
\label{fig::sec1-fig-1}
\end{figure}

Although great strides have been made in automatically delineating \textcolor{black}{cartilages} and \textcolor{black}{bones} \textcolor{black}{\cite{liu2018deep,boutillon2020isbi}}, there is a great need for accurate muscle delineations in managing musculo-skeletal disorders. The task of segmenting muscles from MR images becomes more difficult when the pathology alters the size, shape, texture and global MR appearance of muscles \cite{barnouin2014manual} (Fig.\ref{fig::sec1-fig-1}). Further, the large variability across patients, arising from age-related development and injury, impacts the ability to delineate muscles. To circumvent these difficulties, muscle segmentation is traditionally performed manually\textcolor{black}{,} in a slice-by-slice fashion \cite{tingart2003magnetic}. However, manual segmentation is a time-consuming task and is often imprecise due to intra- and inter-expert variability. Therefore, most musculo-skeletal diagnoses are based on 2D analyses of single images, despite the utility of 3D volume \textcolor{black}{exploration}. Recently, there has been a growing interest in developing \textcolor{black}{automatic} techniques for 3D muscle segmentation, particularly in the area of deploying deep learning methodologies using convolutional encoder-decoders \cite{litjens2017survey}. 

Obstetrical brachial plexus palsy (OBPP), among the most common birth injuries \cite{pons2017shoulder}, is one such pathology in which accurate 3D automatic muscle segmentation could help \textcolor{black}{to} quantify a patient's level of impairment, guide interventional planning \textcolor{black}{or} track treatment progress. OBPP occurs most often during the delivery phase when lateral traction is applied to the head to permit shoulder clearance \cite{oberry2017obstetrical}. It is characterized by the disruption of the peripheral nervous system conducting signals from the spinal cord to shoulders, arms and hands, with an incidence of around 1.4 every 1000 live births \cite{chauhan2014neonatal}. This nerve injury leads to variable muscle denervation, resulting in muscle atrophy with fatty infiltration, growth disruption, muscle atrophy and force imbalances around the shoulder \cite{brochard2014shoulder}. Treatment and prevention of shoulder muscle strength imbalances are main therapeutic goals for children with OBPP who do not fully recover \cite{waters2009correlation}. Patient-specific information related to the degree of muscle atrophy across the shoulder is therefore needed to plan interventions and predict interventional outcomes. Recent work, reporting a clear relationship between muscle atrophy and strength loss for children with OBPP \cite{pons2017shoulder}, demonstrates that an ability to accurately quantify 3D muscle morphology directly translates into an understanding of the force capacity of \textcolor{black}{shoulder} muscles. In this direction, shoulder muscle segmentation on MR images is needed to both quantify individual muscle involvement and analyze shoulder strength balance in children with OBPP. 

Therefore, the purpose of our study is to develop and validate a robust and fully-automated muscle segmentation pipeline, which will support new insights into the evaluation, diagnosis and management of musculo-skeletal diseases. The specific aims are three-fold. First, we aim at studying the feasibility of automatically segmenting pathological shoulder muscle using deep convolutional encoder-decoder networks, based on an available, but small, annotated dataset in children with OBPP \cite{pons2017shoulder}. Second, our work addresses the learning transferability from healthy to pathological data, focusing particularly on how available data from both healthy and pathological shoulder muscles can be jointly exploited for pathological shoulder muscle delineation. Third, extended versions of deep convolutional encoder-decoder architectures, using encoders pre-trained on non-medical data, are investigated to improve the segmentation accuracy. Experiments extend our preliminary results \cite{conze2019deltoid} to four shoulder muscles including deltoid, infraspinatus, supraspinatus and subscapularis. 

\section{Related works}
\label{sec:sec2}

To extract quantitative muscle volume measures, from which forces can be derived \cite{pons2017shoulder}, muscle segmentation is traditionally performed manually in a slice-by-slice \textcolor{black}{manner} \cite{tingart2003magnetic} from MR images. This task is extremely time-consuming and requires tens of minutes to get accurate delineations for one single muscle. Thus, it is not applicable for \textcolor{black}{large} volumes of data typically produced in research studies \textcolor{black}{or} clinical imaging. In addition, manual segmentation is prone to intra- and inter-expert variability, resulting from the irregularity of muscle shapes and the lack of clearly visible boundaries between muscles and surrounding anatomy \cite{pons2018quantifying}. To \textcolor{black}{facilitate} the process, a semi-automatic processing, based on transversal propagations of manually-drawn masks, can be applied \cite{ogier2017individual}. It consists of several ascending and descending non-linear registrations applied to manual masks to finally achieve volumetric results. Although \textcolor{black}{semi-automatic} methods achieve volume segmentation in less time then manual segmentation, they are still time-consuming.

A model-based muscle segmentation incorporating a prior statistical shape model can be employed to delineate muscles boundaries from MR images. A patient-specific 3D geometry is reached based on the deformation of a parametric ellipse fitted to muscle contours, starting from a reduced set of initial slices \cite{sudhoff2009geometry,jolivet2014skeletal}. Segmentation models can be further improved by exploiting \textit{a-priori} knowledge of shape information, relying on internal shape fitting and auto-correction to guide muscle delineation \cite{kim2017automatic}. Baudin et al. \cite{baudin2012prior} combined a statistical shape atlas with a random walks graph-based algorithm to automatically segment individual muscles through iterative linear optimization. Andrews et al. \cite{andrews2015generalized} used a probabilistic shape representation called generalized log-ratio representation that included adjacency information along with a rotationally invariant boundary detector to segment thigh muscles.

Conversely, aligning and merging manually segmented images into specific atlas coordinate spaces can be a reliable alternative to statistical shape models. In this context, various single and multi-atlas methods have been proposed for quadriceps muscle segmentation \cite{ahmad2014atlas,letroter2016volume} relying on non-linear registration. Engstrom et al. \cite{engstrom2011segmentation} used a statistical shape model constrained with probabilistic MR atlases to automatically segment quadratus lumborum. Segmentation of muscle versus fatty tissues has been also performed through possibilistic clustering \cite{barra2002segmentation}, histogram-based thresholding followed by region growing \cite{purushwalkam2013automatic} and active contours \cite{orgiu2016automatic} techniques.

However, all the previously described methods are not perfectly suited for high inter-subject shape variability, significant differences of tissue appearance due to injury and delineations of weak boundaries. Moreover, many of the previously described methods are semi-automatic and hence require prior knowledge, usually associated with high computational costs and large dataset requirements. Therefore, developing a robust fully-automatic muscle segmentation method remains an open and challenging issue, especially when dealing with pathological pediatric data. 

Huge progress has been recently made for automatic image segmentation using deep Convolutional Neural Networks (CNN). Deep CNNs are entirely data-driven supervised learning models formed by multi-layer neural networks \cite{lecun1998gradient}. In contrast to conventional machine learning which requires hand-crafted features and hence specialized knowledge, deep CNNs automatically learn complex hierarchical features directly from data. CNNs obtained outstanding performance for many medical image segmentation tasks \textcolor{black}{\cite{litjens2017survey,tajbakhsh2020embracing}}, which suggests that robust \textcolor{black}{automated delineation} of shoulder muscles from MR images may be achieved using CNN-based segmentation. To our knowledge, no other study has been conducted on shoulder muscle segmentation using deep learning methods.

The simplest way to perform segmentation using deep CNNs consists in classifying each pixel individually by working on patches extracted around them \cite{ciresan2012deep}. Since input patches from neighboring pixels have large overlaps, the same convolutions are computed many time. By replacing fully connected layers with convolutional layers, a Fully Convolutional Network (FCN) can take entire images as inputs and produce likelihood maps instead of single pixel outputs. It removes the need to select representative patches and eliminates redundant calculations due to patch overlaps. In order to avoid outputs with far lower resolution than input shapes, FCNs can be applied to shifted versions of the input images \cite{long2015fully}. Multiple resulting outputs are thus stitched together to get results at full resolution. 

Further improvements can be reached with architectures comprising a regular FCN to extract features and capture context, followed by an up-sampling part that enables to recover the input resolution using up-convolutions \cite{litjens2017survey}. Compared to patch-based or \textit{shift-and-stitch} methods, it allows a precise localization in a single pass while taking into account the full image context. Such architecture made of paired networks is called Convolutional Encoder-Decoder (CED). 

U-Net \cite{ronneberger2015unet} is the most well-known CED in the medical image analysis community. It has a symmetrical architecture with equal amount of down-sampling and up-sampling layers between contracting and expanding paths (Fig.\ref{fig::sec3-3-fig-1}\textit{a}). The encoder gradually reduces the spatial dimension with pooling layers whereas the decoder gradually recovers object details and spatial dimension. One key aspect of U-Net is the use of \textcolor{black}{shortcuts (so-called \textit{skip connections})} which concatenate features from the encoder to the decoder to help in recovering object details while improving localization accuracy. By allowing information to \textcolor{black}{directly} flow from low-level to high-level feature maps, faster convergence is achieved. This architecture can be exploited for 3D volume segmentation \cite{cciccek2016unet} by replacing all 2D operations with their 3D counterparts but at the cost of computational speed and memory consumption. Processing 2D slices independently before reconstructing 3D medical volumes remains a simpler alternative. Instead of cross-entropy used as loss function, the extension of U-Net proposed in \cite{milletari2016vnet} directly minimizes a segmentation error to handle class imbalance between foreground and background.

\section{Material and methods}
\label{sec:sec3}

In this work, we develop and validate a fully-automatic methodology for pathological shoulder muscle segmentation through deep CEDs (Sect.\ref{sec:sec2}), using a pediatric OPBB dataset (Sect.\ref{sec:sec3-1}). Healthy versus pathological learning transferability is addressed in Sect.\ref{sec:sec3-2}. Extended deep CED architectures with pre-trained encoders are proposed in Sect.\ref{sec:sec3-3}. Assessment is performed using dedicated \textcolor{black}{evaluation} metrics (Sect.\ref{sec:sec3-4}).

\subsection{Imaging dataset}
\label{sec:sec3-1}

Data collected from a previous study \cite{pons2017shoulder} investigating the muscle volume-strength relationship in 12 children with unilateral OPBB (averaged age \textcolor{black}{of} $12.1\pm3.3$ years) formed the basis of the current study. In this IRB approved study, informed consents from a legal guardian and assents from the participants were obtained for all subjects. If a participant was over 18 years of age, only informed consent was obtained from that participant. For each patient, two 3D axial-plane T1-weighted gradient-echo MR images were acquired: one for the affected shoulder and another for the unaffected \textcolor{black}{one}. For each  image set, equally spaced  2D axial slices were selected for four different  rotator cuff  muscles:  deltoid, infraspinatus, supraspinatus and subscapularis. These slices were annotated by an expert in pediatric physical medicine and rehabilitation to reach pixel-wise groundtruth delineations. Image size for axial slices are constant for each subject ($416\hspace{0.02cm}\times\hspace{0.02cm}312$ pixels). Image resolution varies from $0.55\hspace{-0.05cm}\times\hspace{-0.03cm}0.55$ to $0.63\hspace{-0.02cm}\times\hspace{-0.02cm}0.63$mm, allowing a finer resolution for smaller subjects. The number of axial slices fluctuates from $192$ to $224$, whereas slice thickness remains unchanged ($1.2$mm). Overall, we had 374 (resp. 395) annotated axial slices for deltoid, 306 (367) for infraspinatus, 238 (208) for supraspinatus and 388 (401) for subscapularis across 2400 (2448) axial slices arising from 12 affected (unaffected) shoulders. Among these 24 MR image sets, pairings between affected and unaffected shoulders are known. Due to sparse annotations (Fig.\ref{fig::sec1-fig-1}), deep CEDs exploit as inputs 2D axial slices and produce 2D segmentation masks which can be then stacked to recover a 3D volume for clinical purposes. Among the images from the affected side, $8$ are from right shoulders (\texttt{R-P-}\textcolor{black}{$\{$}\texttt{0134},\texttt{0684},\texttt{0382},\texttt{0447},\texttt{0660},\texttt{0737},\texttt{0667},\texttt{0277}\textcolor{black}{$\}$}) whereas $4$ correspond to left shoulders (\texttt{L-P-}\textcolor{black}{$\{$}\texttt{0103},\texttt{0351},\texttt{0922},\texttt{0773}\textcolor{black}{$\}$}). Training images displaying a right (left) shoulder are flipped when a left (right) shoulder is considered for test.

\subsection{Healthy versus pathological learning transferability}
\label{sec:sec3-2}

\begin{figure}
\vspace{-0.2cm} 
\begin{tabular}{c}
\centering \includegraphics[height=5.8cm]{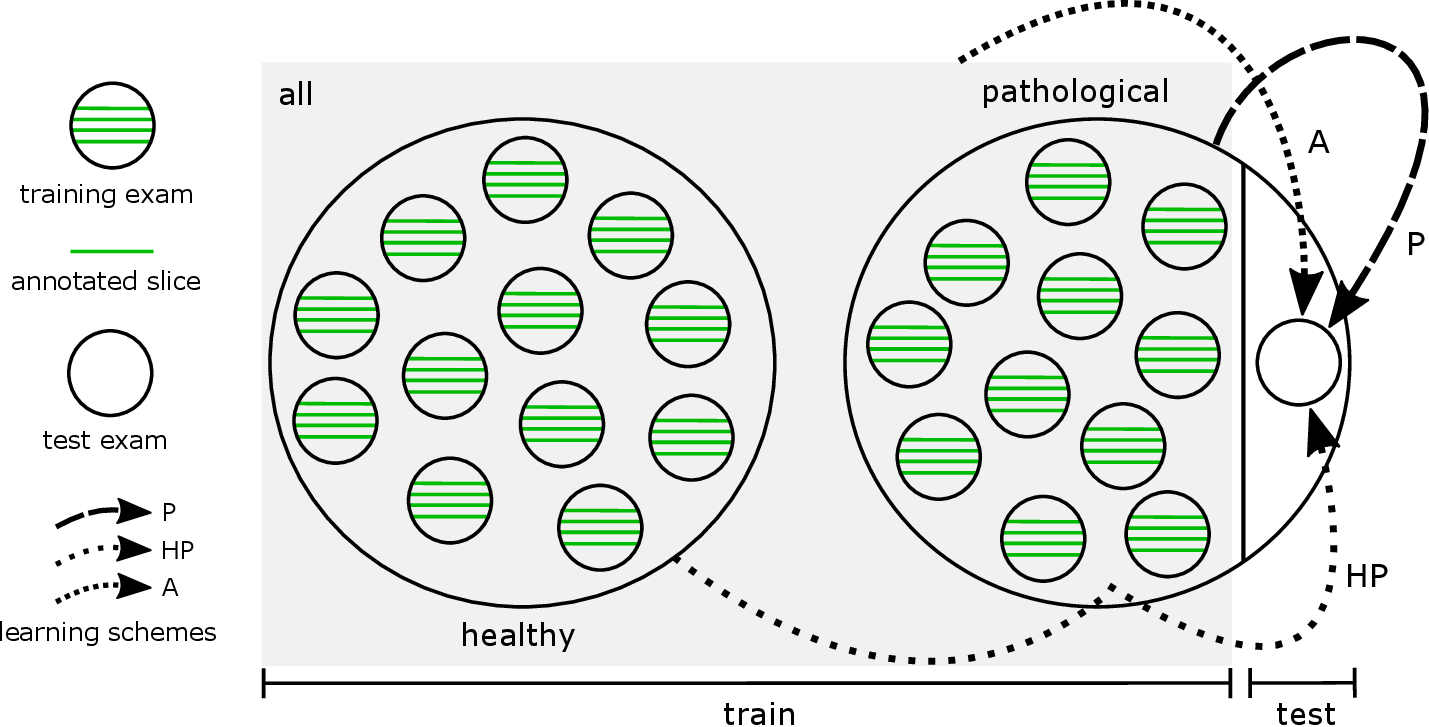} \cr
\end{tabular} 
\caption{Three different learning schemes (\texttt{P}, \texttt{HP}, \texttt{A}) involved in a leave-one-out setting for deep learning-based pathological shoulder muscle segmentation.}
\label{fig::sec3-2-fig-1}
\end{figure}

In the context of OBPP, the limited availability of both healthy and pathological data for image segmentation brings new queries related to the learning transferability from healthy to pathological structures. This aspect is particularly suitable to musculo-skeletal pathologies for two reasons. First, despite different shapes and sizes due to growth and atrophy, healthy and pathological muscles may share common characteristics such as anatomic locations and overall aspects. Second, combining healthy and pathological data for deep learning-based segmentation can  act as a smart data augmentation strategy when faced with limited annoted data. In exploring the combined use of healthy and pathological data for pathological muscle segmentation, determining the optimal learning scheme is crucial. Thus, three different learning schemes (Fig.\ref{fig::sec3-2-fig-1}) employed with deep CEDs are considered:

\begin{itemize}
\item[-] \textbf{pathological only} (\texttt{P}): the most common configuration consists in exploiting groundtruth annotations made on impaired shoulder muscles only, \textcolor{black}{making the hypothesis} that CED features extracted from healthy examinations are not suited enough for pathological anatomies. 
\item[-] \textbf{healthy transfer to pathological} (\texttt{HP}): another strategy deals with transfer learning and fine tuning from healthy to pathological muscles. In this context, a first CED is trained using groundtruth segmentations from unaffected shoulders only. The weights of the resulting model are then used as initialization for a second CED network which is trained using pathological inputs only. 
\item[-] \textbf{simultaneous healthy and pathological} (\texttt{A}\textcolor{black}{\footnotemark[1]}): the last configuration consists in training a CED with a groundtruth dataset comprising annotations made on both healthy and pathological shoulder muscles, which allows to benefit from a more consequent dataset.
\end{itemize}

\footnotetext[1]{\textcolor{black}{\texttt{A} stands for `all'}}

By comparing these different training strategies, we evaluate the benefits brought by combining healthy and pathological data together in terms of model generalizability. The balance between data augmentation and healthy versus pathological muscle variability is a crucial question which has never been investigated for muscle segmentation. These three different schemes, referred as \texttt{P} (pathological only), \texttt{HP} (healthy transfer to pathological) and \texttt{A} (simultaneous healthy and pathological) are compared in a leave-one-out \textcolor{black}{fashion} (Fig.\ref{fig::sec3-2-fig-1}). The overall dataset is divided into healthy and pathological MR examinations. Iteratively, one pathological examination is extracted from the pathological dataset and considered as test examination for muscle segmentation. To avoid any \textcolor{black}{bias} for \texttt{HP} and \texttt{A}, annotated data from the healthy shoulder of the patient whose pathological shoulder is considered for test is not used during training.

For all schemes, deep CED networks are trained using data augmentation since the amount of available training data is limited. \textcolor{black}{Training 2D axial slices undergo random scaling, rotation, shearing and shifting on both directions to teach the network the desired invariance and robustness properties \cite{ronneberger2015unet}. In practice,} 100 augmented images are produced for one single training axial slice. Comparisons between \texttt{P}, \texttt{HP} and \texttt{A} schemes are performed using standard U-Net \cite{ronneberger2015unet} with 10 epochs, a batch size of 10 images, an \textit{Adam} optimizer with $10^{-4}$ as learning rate for stochastic optimization, a fuzzy Dice score as loss function and randomly initialized weights for convolutional filters. Models were implemented using Keras and trained \textcolor{black}{with} a single Nvidia GeForce GTX 1080 Ti GPU with $11$Gb/s. Once training is performed, predictions for one single axial slice take 28ms only which is suitable for routine clinical practice.

\subsection{Extended architectures with pre-trained encoders}
\label{sec:sec3-3}

Contrary to deep classification networks which are usually pre-trained on a very large image dataset, CED architectures used for segmentation are typically trained from scratch, relying on randomly initialized weights. Reaching a generic model without over-fitting is therefore \textcolor{black}{challenging}, especially when only a small amount of images is available. As suggested in \cite{iglovikov2018ternausnet}, the encoder part of a deep CED network can be replaced by a well-known classification network whose weights are pre-trained on an initial classification task. It allows to exploit transfer learning from large datasets such as ImageNet \cite{russakovsky2015imagenet} for deep learning-based segmentation. In the literature, the encoder part of a deep CED has been already replaced by pre-trained VGG-11 \cite{iglovikov2018ternausnet} and ABN WideResnet-38 \cite{iglovikov2018ternausnetv2} with improvements compared to their randomly weighted counterparts.

Following this idea, we propose to extend the standard U-Net architecture (Sect.\ref{sec:sec2}) by exploiting another simple network from the VGG family \cite{simonyan2014very} as encoder, namely the VGG-16 architecture. To improve performance, this encoder branch is pre-trained on ImageNet \cite{russakovsky2015imagenet}. This database has been designed for object recognition purposes and contains more than $1$ million natural images from 1000 classes. Pre-training our deep CED dedicated to muscle image segmentation using non-medical data is an efficient way to reduce the data scarcity issue while improving model generalizability \cite{yosinski2014transferable}. Pre-trained models can not only improve predictive performance but also require less training time to reach convergence for the target task. In particular, low-level features captured by first convolutional layers are usually shared between different image types which explains the success of transfer learning between tasks.

\begin{figure}
\hspace{-2.9cm} \begin{tabular}{cc}
\rotatebox{90}{\small \textcolor{white}{----} a) \texttt{U-Net} \cite{ronneberger2015unet}} & 
\includegraphics[width=16.6cm]{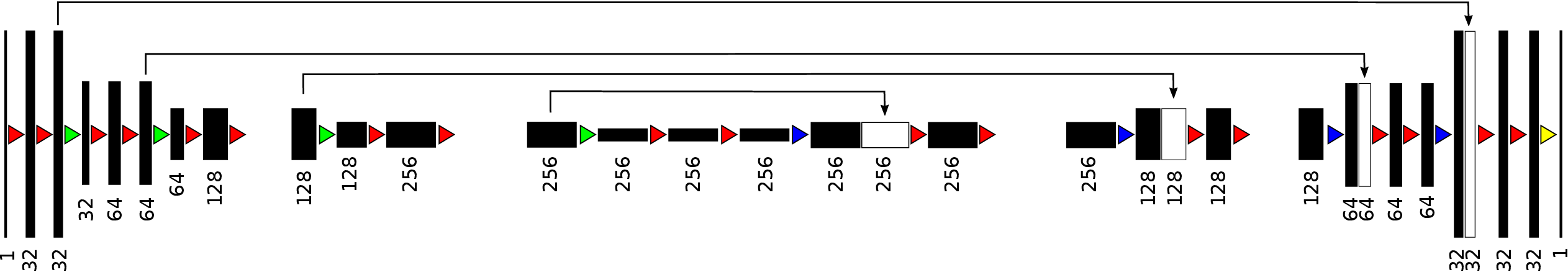} \vspace{0.2cm} \cr
\rotatebox{90}{\small \textcolor{white}{--.} b) \texttt{v16pU-Net}} & 
\includegraphics[width=16.6cm]{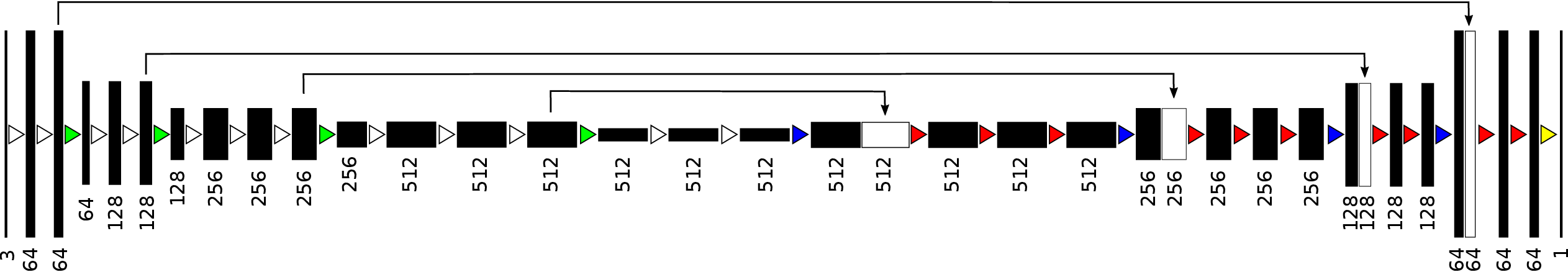} \vspace{0.2cm} \cr
&
\includegraphics[width=15cm]{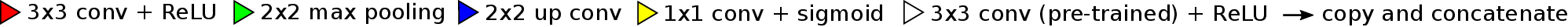} \cr
\end{tabular} 
\caption{Extension of U-Net \cite{ronneberger2015unet} by exploiting as encoder a slightly modified VGG-16 \cite{simonyan2014very} with weights pre-trained on ImageNet \cite{russakovsky2015imagenet}, following \cite{iglovikov2018ternausnet,iglovikov2018ternausnetv2}. The decoder is modified to get an exactly symmetrical construction while keeping \textit{skip connections}.}
\label{fig::sec3-3-fig-1}
\end{figure}

The VGG-16 encoder (Fig.\ref{fig::sec3-3-fig-1}\textit{b}) consists of sequential layers including $3\hspace{-0.05cm}\times\hspace{-0.05cm}3$ convolutional layers followed by Rectified Linear Unit (ReLU) activation functions. Reducing the spatial size of the representation is handled by $2\hspace{-0.05cm}\times\hspace{-0.05cm}2$ max pooling layers. Compared to standard U-Net (Fig.\ref{fig::sec3-3-fig-1}\textit{a}), the first convolutional layer generates 64 channels instead of $32$. As the network deepens, the number of channels doubles after each max pooling until it reaches 512 (256 for classical U-Net). After the second max pooling operation, the number of convolutional layers differ from U-Net with patterns of $3$ consecutive convolutional layers instead of $2$, following the original VGG-16 architecture. In addition, input images are extended from one single greyscale channel to 3 channels by repeating the same content in order to respect the dimensions of the RGB ImageNet images used for encoder pre-training. The only differences with VGG-16 rely in the fact that the last convolutional layer as well as top layers including fully-connected layers and softmax have been omitted. The two last convolutional layers taken from VGG-16 serve as central part of the CED and separate both contracting and expanding paths.

The extension of the U-Net encoder is transferred to the decoder branch by adding $2$ convolutional layers as well as more \textcolor{black}{feature} channels to get an exactly symmetrical construction while keeping \textit{skip connections}. Contrary to encoder weights which are initialized using pre-training performed on ImageNet, decoder weights are set randomly. As for U-Net, a final $1\hspace{-0.05cm}\times\hspace{-0.05cm}1$ convolutional layer followed by a sigmoid activation function achieves pixel-wise segmentation masks whose resolution is the same as input slices.

Pathological shoulder muscle segmentation using the standard U-Net architecture \cite{ronneberger2015unet} as well as the proposed extension without (\texttt{v16U-Net}) and with (\texttt{v16pU-Net}) weights pre-trained on ImageNet is performed through leave-one-out experiments. In this context, we rely on training scheme \texttt{A} combining both healthy and pathological data (Sect.\ref{sec:sec3-2}). As previously, networks are trained with data augmentation, 10 epochs, a batch size of 10 images, \textcolor{black}{an} \textit{Adam} optimizer and a fuzzy Dice score used as loss function. Learning rates change from U-Net and \texttt{v16pU-Net} ($10^{-4}$) to \texttt{v16U-Net} ($5\times10^{-5}$) to avoid divergence for deep networks trained with randomly selected weights.

\subsection{Segmentation assessment}
\label{sec:sec3-4}

To assess both healthy versus pathological learning transferability (Sect.\ref{sec:sec3-2}) and extended pre-trained deep convolutional architectures (Sect.\ref{sec:sec3-3}), the accuracy of automatic pathological shoulder muscle segmentation is quantified based on Dice ($\frac{2TP}{2TP+FP+FN}$), sensitivity ($\frac{TP}{TP+FN}$), specificity ($\frac{TN}{TN+FP}$) and Jaccard ($\frac{TP}{TP+FP+FN}$) scores \textcolor{black}{(in \%)} where TP, FP, TN and FN are the number of true or false positive and negative pixels. \textcolor{black}{Evaluations also rely on the Cohen's kappa coefficient ($\frac{p_o-p_e}{1-p_e}$) in \% where $p_o$ and $p_e$ are the relative observed agreement and the hypothetical probability of chance agreement. In practice, $p_o=\frac{TP+TN}{TP+FN+FP+TN}$ which corresponds to the accuracy and $p_e=\frac{(TP+FN)\times(TP+FP)}{TP+FN+FP+TN}+\frac{(FP+TN)\times(FN+TN)}{TP+FN+FP+TN}$. Finally, we} exploit an absolute surface estimation error (\texttt{ASE}) which compares groundtruth and estimated muscle surfaces defined in mm$^2$ from segmentation masks. These scores tend to provide a complete assessment of the ability of CED models to provide contours identical to those manually performed. \textcolor{black}{Reported results} are averaged among all annotated slices arising from the 12 pathological shoulder examinations. \textcolor{black}{Network parameters are those reaching the best fuzzy Dice test scores during training}.

\section{Results and discussion}
\label{sec:sec4}

\subsection{Healthy versus pathological learning transferability}
\label{sec:sec4-1}

\begin{figure*}
\hspace{-3cm} \begin{tabular}{cccc}
\includegraphics[height=4.4cm]{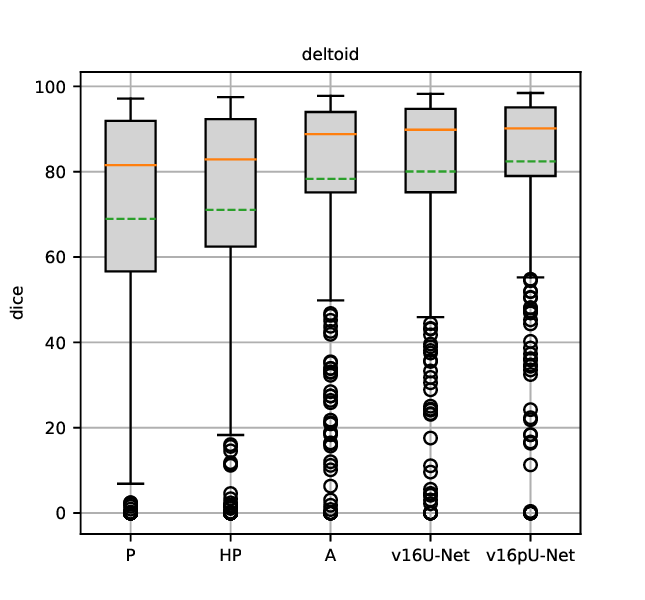} &
\hspace{-0.8cm} \includegraphics[height=4.4cm]{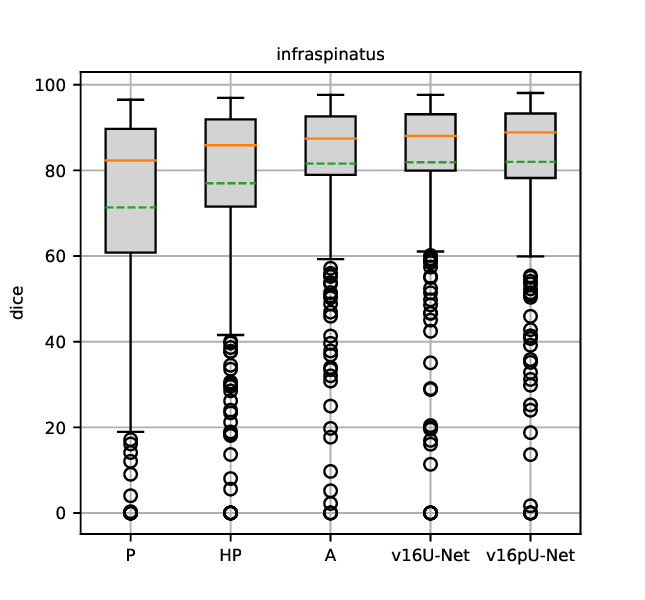} &
\hspace{-0.8cm} \includegraphics[height=4.4cm]{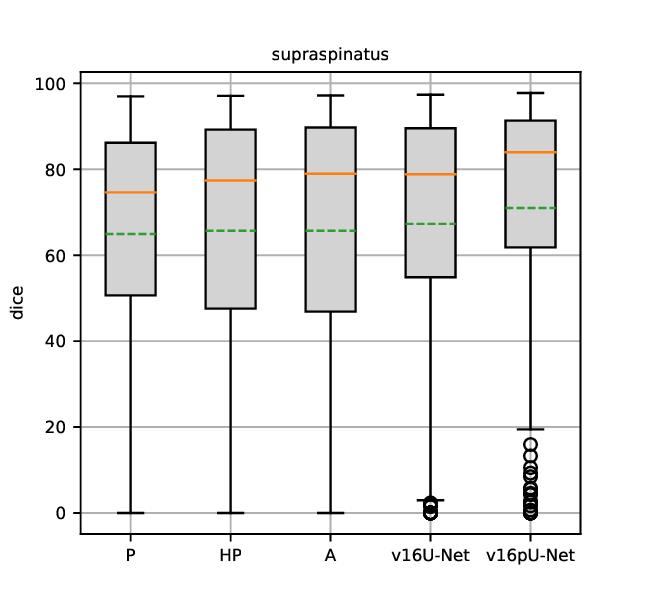} &
\hspace{-0.8cm} \includegraphics[height=4.4cm]{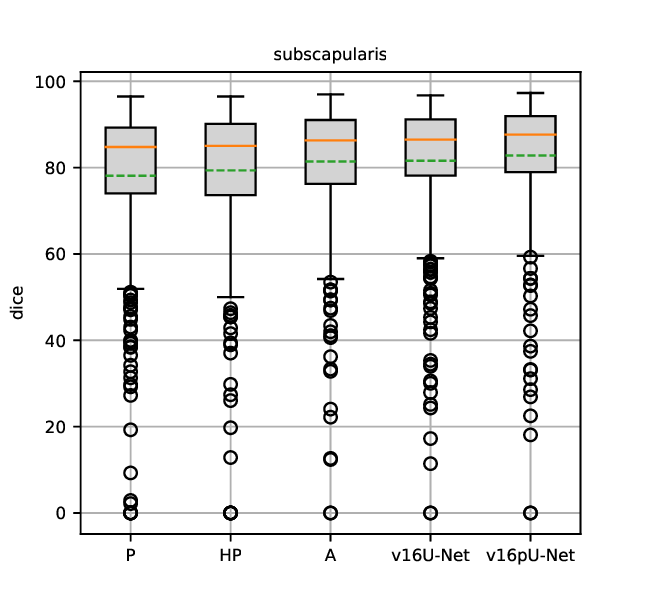} \vspace{-0.4cm} \cr
\includegraphics[height=4.4cm]{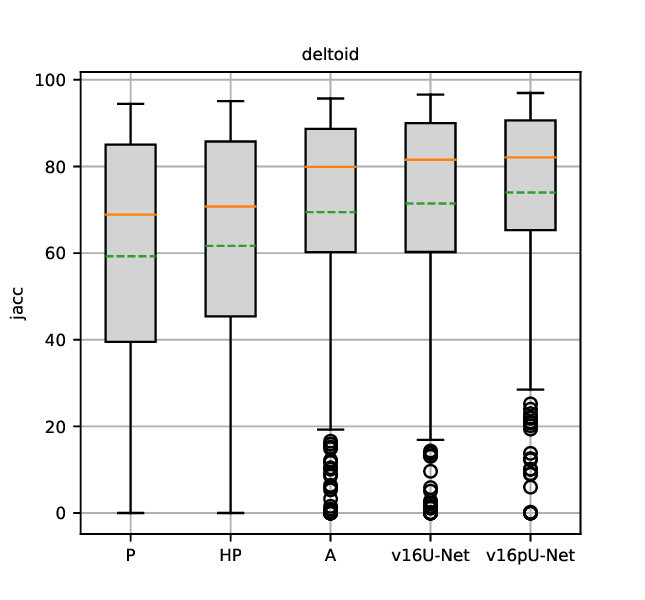} &
\hspace{-0.8cm} \includegraphics[height=4.4cm]{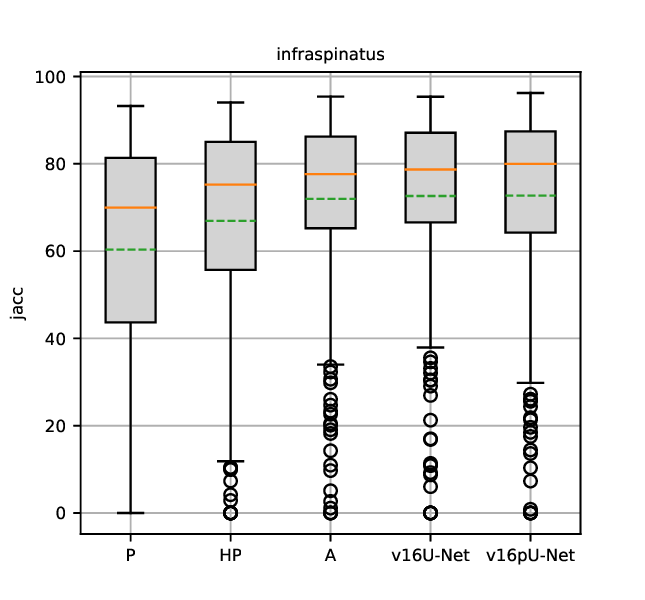} &
\hspace{-0.8cm} \includegraphics[height=4.4cm]{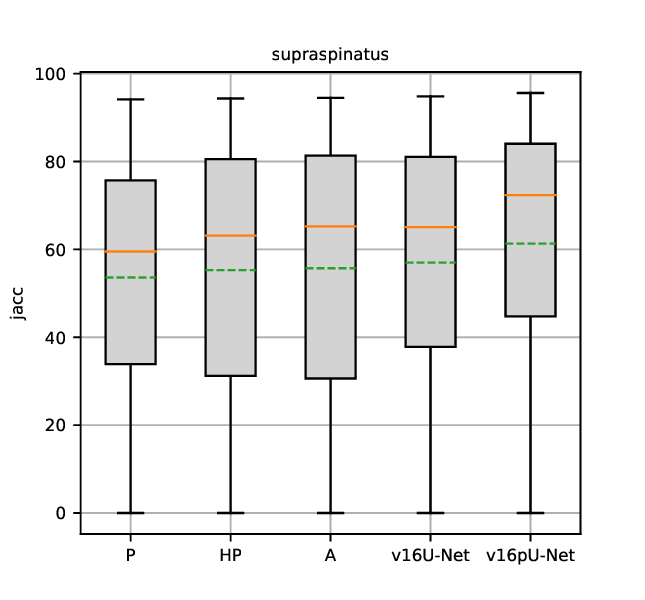} &
\hspace{-0.8cm} \includegraphics[height=4.4cm]{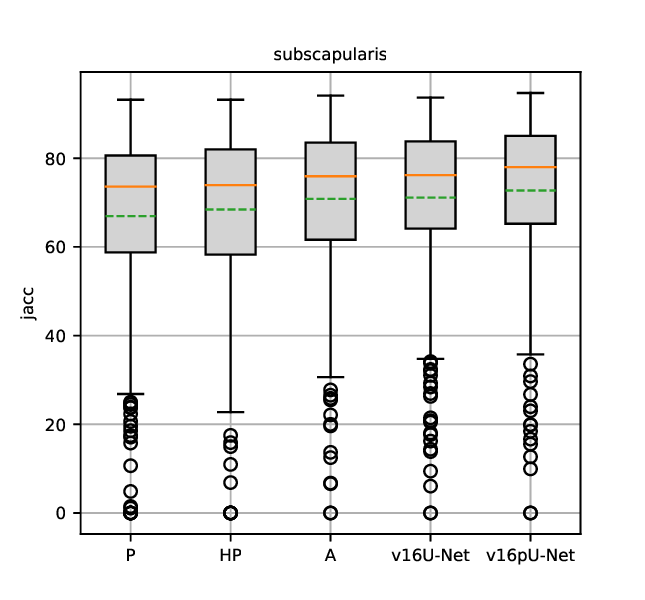} \vspace{-0.4cm} \cr
\includegraphics[height=4.4cm]{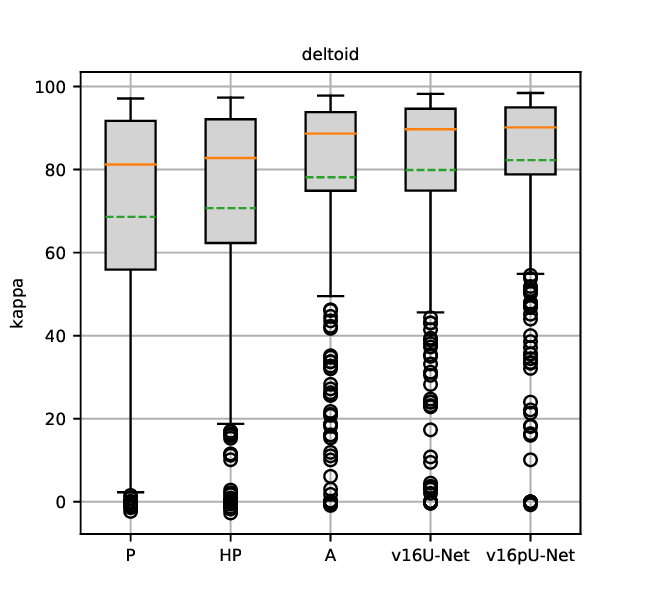} &
\hspace{-0.8cm} \includegraphics[height=4.4cm]{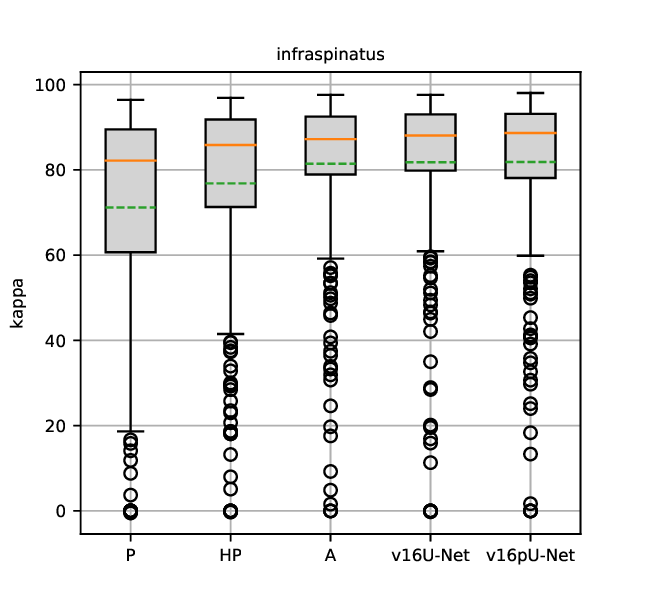} &
\hspace{-0.8cm} \includegraphics[height=4.4cm]{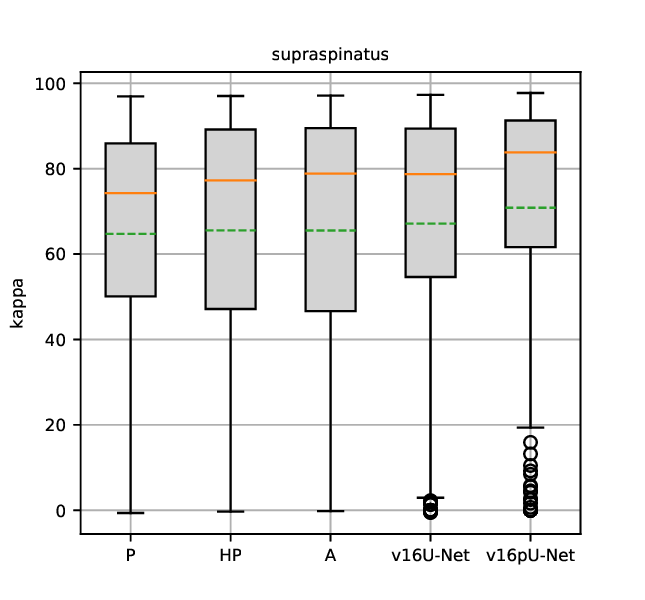} &
\hspace{-0.8cm} \includegraphics[height=4.4cm]{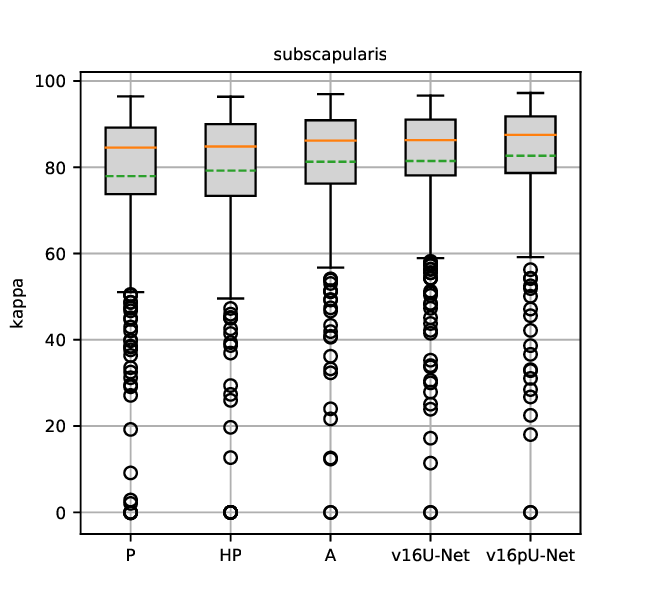} \vspace{-0.4cm} \cr
\includegraphics[height=4.4cm]{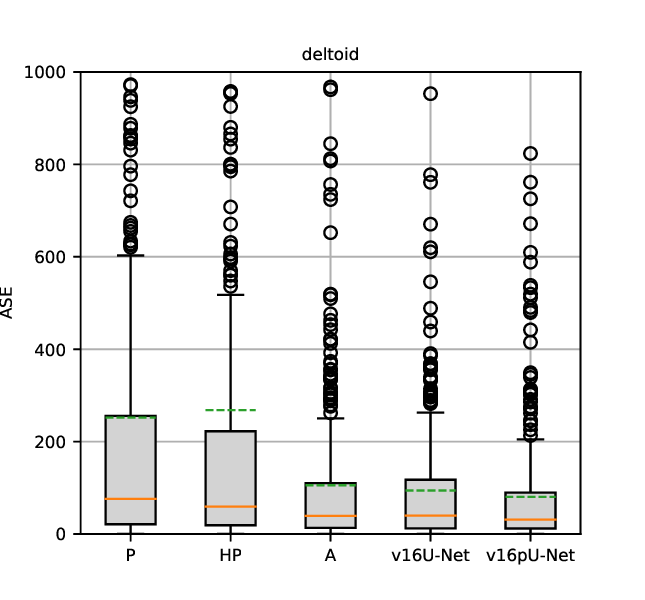} &
\hspace{-0.8cm} \includegraphics[height=4.4cm]{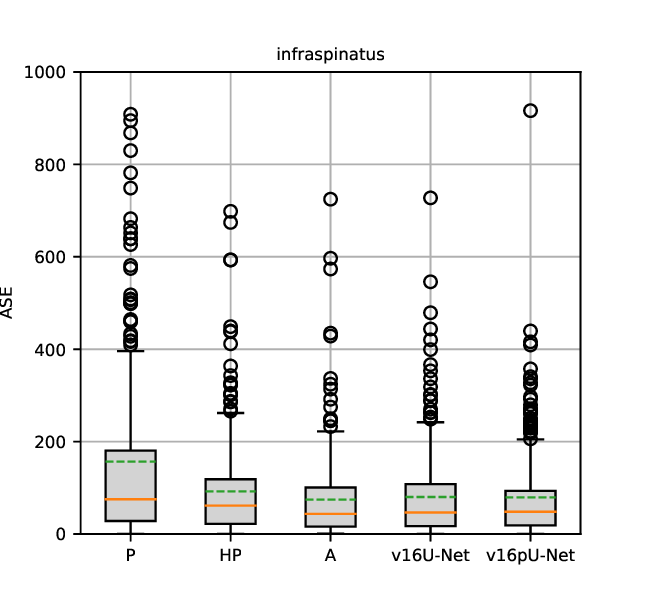} &
\hspace{-0.8cm} \includegraphics[height=4.4cm]{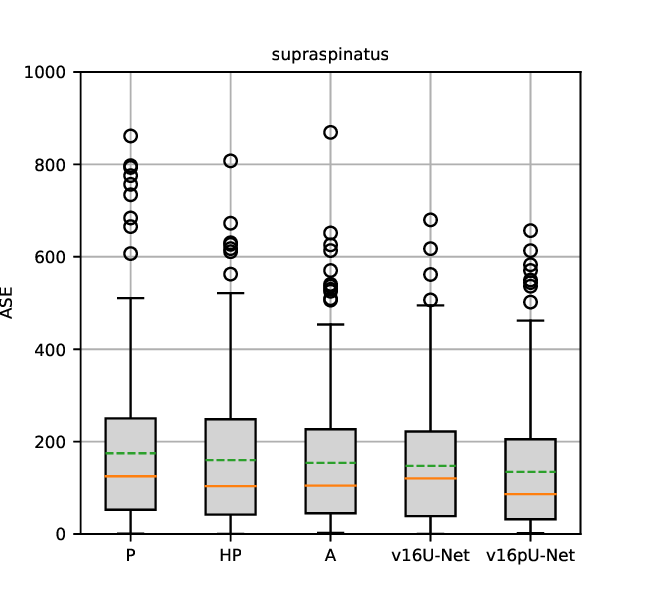} &
\hspace{-0.8cm} \includegraphics[height=4.4cm]{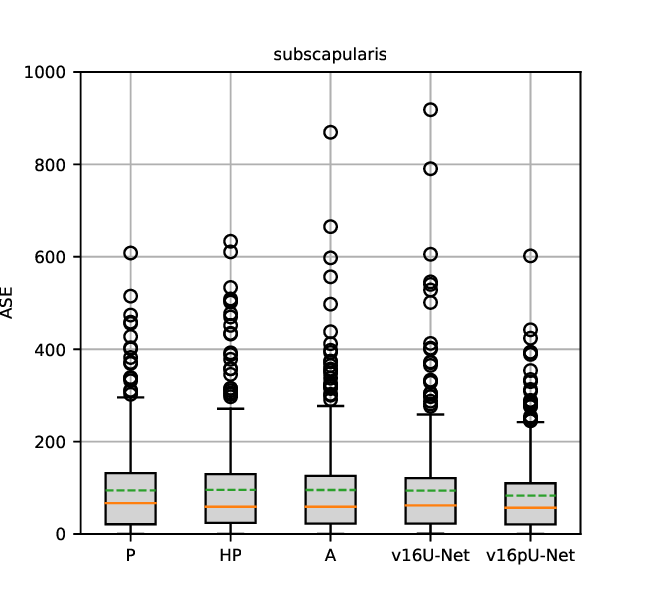} \vspace{-0.1cm} \cr
\end{tabular} 
\vspace{-0.3cm}
\caption{\textcolor{black}{Box plots on Dice, Jaccard, Cohen's kappa and absolute surface error (\texttt{ASE}) slice-wise scores over the pathological dataset using convolutional encoder-decoders (U-Net \cite{ronneberger2015unet}, \texttt{v16U-Net}, \texttt{v16pU-Net}) embedded with learning schemes \texttt{P}, \texttt{HP} and \texttt{A}. Dashed green and solid orange lines respectively represent means and medians.}}
\label{fig::sec4-1-fig-1}
\end{figure*}

The highest performance is achieved when both healthy and pathological data are simultaneously used for training (\texttt{A}), with Dice scores of $78.32\%$ for deltoid, $81.58\%$ for infraspinatus and $81.41\%$ for subscapularis (Tab.\ref{tab::sec4-tab-1}). Scheme \texttt{A} outperforms transfer learning and fine tuning (\texttt{HP}) from $4$ to $7\%$ in terms of Dice. However, this conclusion does not apply to supraspinatus for which \texttt{A} and \texttt{HP} schemes achieve the same performance in Dice ($\approx\hspace{-0.02cm}65.7\%$) \textcolor{black}{and Cohen's kappa ($\approx\hspace{-0.02cm}65.6\%$)}. In particular, \texttt{A} increases the sensitivity ($65.55\%$ instead of $63.16\%$) but provides a slightly smaller specificity, compared to \texttt{HP}. \textcolor{black}{In this specific case, medians are nevertheless rather in favour of \texttt{A} compared to means (Fig.\ref{fig::sec4-1-fig-1}).} Comparing \texttt{ASE} from \texttt{HP} to \texttt{A} reveals improvements for all shoulder muscles, including deltoid whose surface estimation error decreases from $268$ to $105.5$mm$^2$. The same finding arises when studying Jaccard scores whose gains are $7.8\%$ and $6.5\%$ for deltoid and subscapularis. \textcolor{black}{The Cohen's kappa coefficient jumps from $70.73\%$ ($76.85\%$) to $78.15\%$ ($81.45\%$) for deltoid (infraspinatus)}. Therefore, directly combining healthy and pathological data appears a better strategy than dividing training into two parts, focusing on first healthy and then pathological data via transfer learning. Further, exploiting annotations for the pathological shoulder muscles only (\texttt{P}) is the worst training strategy (Tab.\ref{tab::sec4-tab-1}, \textcolor{black}{Fig.\ref{fig::sec4-1-fig-1}}), especially for deltoid (Dice loss of $10\%$ from \texttt{A} to \texttt{P}). However, results for subscapularis deviate from this result, with higher similarity scores \textcolor{black}{(except for kappa)} compared to \texttt{HP} combined with the best \texttt{ASE} ($94.56$mm$^2$). \textcolor{black}{In} general, the CED features extracted from healthy examinations are suited enough for pathological anatomies while acting as an efficient data augmentation strategy. 

\begin{table*}
\small 
\hspace{-1.48cm} \begin{tabular}{|c|l|c|c|c?c|c|}
\hline
\multirow{2}{*}{metric} & \multicolumn{1}{c|}{scheme} & \texttt{P} & \texttt{HP} & \multicolumn{3}{c|}{\texttt{A}} \\ \cline{2-7} 
                        & \multicolumn{1}{c|}{network} & \multicolumn{3}{c?}{\texttt{U-Net} \cite{ronneberger2015unet}} & \texttt{v16U-Net} & \texttt{v16pU-Net} \\ \hline \hline
                        
\multirow{4}{*}{\rotatebox{90}{\texttt{dice} \textcolor{black}{$\uparrow$}}}  & deltoid & 68.94$\pm$29.9 & 71.05$\pm$29.5 & \textit{\underline{78.32}}$\pm$24.4 & 80.05$\pm$23.1 & \textbf{82.42}$\pm$20.4 \\ \cline{2-7} 
& infraspinatus & 71.38$\pm$24.7 & 77.00$\pm$22.5 & \textit{\underline{81.58}}$\pm$18.3 & 81.91$\pm$19.0 & \textbf{81.98}$\pm$18.6 \\ \cline{2-7} 
& supraspinatus & 64.94$\pm$28.0 & \textit{\underline{65.69}}$\pm$29.6 & 65.68$\pm$30.7 & 67.30$\pm$29.4 & \textbf{70.98}$\pm$28.7 \\ \cline{2-7} 
& subscapularis & 78.10$\pm$18.1 & 74.55$\pm$25.2 & \textit{\underline{81.41}}$\pm$15.0 & 81.58$\pm$15.2 & \textbf{82.80}$\pm$14.4 \\ \hline \hline
\multirow{4}{*}{\rotatebox{90}{\texttt{sens} \textcolor{black}{$\uparrow$}}} & deltoid & 70.85$\pm$30.5 & 70.74$\pm$29.5 & \textit{\underline{78.92}}$\pm$25.4 & 81.45$\pm$23.7 & \textbf{83.80}$\pm$21.3 \\ \cline{2-7} 
& infraspinatus & 72.12$\pm$26.4 & 79.45$\pm$23.1 & \textit{\textbf{\underline{84.61}}}$\pm$18.2 & 83.74$\pm$18.6 & 83.48$\pm$19.0 \\ \cline{2-7} 
& supraspinatus & 64.02$\pm$31.8 & 63.16$\pm$33.2 & \textit{\underline{65.55}}$\pm$34.5 & 67.21$\pm$33.0 & \textbf{68.60}$\pm$32.3 \\ \cline{2-7} 
& subscapularis & 78.89$\pm$19.7 & 74.75$\pm$27.3 & \textit{\underline{82.53}}$\pm$18.1 & 81.75$\pm$18.8 & \textbf{84.36}$\pm$16.5 \\ \hline \hline
\multirow{4}{*}{\rotatebox{90}{\texttt{spec} \textcolor{black}{$\uparrow$}}} & deltoid & 99.61$\pm$0.80 & 99.56$\pm$1.07 & \textit{\textbf{\underline{99.85}}}$\pm$0.19& 99.82$\pm$0.22 & 99.84$\pm$0.22 \\ \cline{2-7} 
& infraspinatus & 99.82$\pm$0.23 & 99.82$\pm$0.22 & \textit{\underline{99.84}}$\pm$0.18 & \textbf{99.86}$\pm$0.17 & \textbf{99.86}$\pm$0.18 \\ \cline{2-7} 
& supraspinatus & 99.86$\pm$0.18 & \textit{\underline{99.90}}$\pm$0.13 & 99.88$\pm$0.15 & 99.86$\pm$0.17 & \textbf{99.91}$\pm$0.12 \\ \cline{2-7} 
& subscapularis & 99.86$\pm$0.13 & 99.83$\pm$0.28 & \textit{\underline{99.87}}$\pm$0.13 & \textbf{99.88}$\pm$0.12 & 99.86$\pm$0.15 \\ \hline \hline
\multirow{4}{*}{\rotatebox{90}{\texttt{jacc} \textcolor{black}{$\uparrow$}}} & deltoid & 59.27$\pm$29.7 & 61.68$\pm$29.3 & \textit{\underline{69.48}}$\pm$26.0 & 71.46$\pm$24.9 & \textbf{74.00}$\pm$22.8 \\ \cline{2-7} 
& infraspinatus & 60.32$\pm$25.6 & 66.91$\pm$24.0 & \textit{\underline{72.00}}$\pm$20.4 & 72.63$\pm$20.6 & \textbf{72.71}$\pm$21.0 \\ \cline{2-7} 
& supraspinatus & 53.61$\pm$27.1 & 55.27$\pm$29.3 & \textit{\underline{55.70}}$\pm$30.1 & 56.98$\pm$28.7 & \textbf{61.31}$\pm$28.7 \\ \cline{2-7} 
& subscapularis & 66.93$\pm$19.6 & 64.31$\pm$24.7 & \textit{\underline{70.83}}$\pm$17.6 & 71.13$\pm$17.7 & \textbf{72.72}$\pm$17.16 \\ \hline \hline
\multirow{4}{*}{\rotatebox{90}{\textcolor{black}{\texttt{kappa}} \textcolor{black}{$\uparrow$}}} & deltoid & \textcolor{black}{68.63$\pm$30.0} & \textcolor{black}{70.73$\pm$29.7} & \textcolor{black}{\textit{\underline{78.15}}$\pm$24.4} & \textcolor{black}{79.89$\pm$23.2} & \textcolor{black}{\textbf{82.28}$\pm$20.5} \\ \cline{2-7} 
& infraspinatus & \textcolor{black}{71.19$\pm$24.7} & \textcolor{black}{76.85$\pm$22.5} & \textcolor{black}{\textit{\underline{81.45}}$\pm$18.3} & \textcolor{black}{81.79$\pm$19.0} & \textcolor{black}{\textbf{81.86}$\pm$18.7} \\ \cline{2-7} 
& supraspinatus & \textcolor{black}{64.76$\pm$28.0} & \textcolor{black}{\textit{\underline{65.56}}$\pm$29.6} & \textcolor{black}{65.55$\pm$30.7} & \textcolor{black}{67.16$\pm$29.4} & \textcolor{black}{\textbf{70.87}$\pm$28.7} \\ \cline{2-7}  
& subscapularis & \textcolor{black}{77.95$\pm$18.1} & \textcolor{black}{79.23$\pm$17.1} & \textcolor{black}{\textit{\underline{81.27}}$\pm$15.0} & \textcolor{black}{81.45$\pm$15.2} & \textcolor{black}{\textbf{82.67}$\pm$14.4} \\ \hline \hline
\multirow{4}{*}{\rotatebox{90}{\texttt{ASE} \textcolor{black}{$\downarrow$}}} & deltoid & 252.0$\pm$421.6 & 268.0$\pm$507.8 & \textit{\underline{105.5}}$\pm$178.9 & 94.23$\pm$139.2 & \textbf{80.38}$\pm$127.5 \\ \cline{2-7} 
& infraspinatus & 156.8$\pm$228.7 & 92.37$\pm$105.9 & \textit{\textbf{\underline{74.47}}}$\pm$92.8 & 80.11$\pm$96.2 & 79.17$\pm$96.9 \\ \cline{2-7} 
& supraspinatus & 174.8$\pm$164.0 & 159.9$\pm$153.5 & \textit{\underline{153.9}}$\pm$146.0 & 147.5$\pm$129.4 & \textbf{134.6}$\pm$135.5 \\ \cline{2-7} 
& subscapularis & \textit{\underline{94.56}}$\pm$95.5 & 102.0$\pm$110.7 & 95.19$\pm$109.0 & 94.06$\pm$111.3 & \textbf{82.95}$\pm$86.88 \\ \hline
\end{tabular}
\caption{Quantitative assessment of \textcolor{black}{convolutional encoder-decoders} (U-Net \cite{ronneberger2015unet}, \texttt{v16U-Net}, \texttt{v16pU-Net}) embedded with learning schemes \texttt{P}, \texttt{HP} and \texttt{A} over the pathological dataset in Dice, \textcolor{black}{sensitivity}, specificity, Jaccard, \textcolor{black}{Cohen's kappa} (\%) as well as absolute surface error \textcolor{black}{(mm$^{2}$)}. Best results are in bold. Italic underlined scores highlight best results among learning schemes employed with U-Net.}
\label{tab::sec4-tab-1}
\end{table*}

Accuracy scores for supraspinatus are globally worse than for other muscles \textcolor{black}{(Fig.\ref{fig::sec4-1-fig-1})} since its thin and elongated shape can strongly vary across patients \cite{kim2017automatic}. Moreover, we notice the presence of a single severely atrophied supraspinatus (\texttt{L-P-0922}) among the set of pathological examinations. Dice results for this single muscle is $42.99\%$ for \texttt{P} against $38.59\%$ and $32.33\%$ for \texttt{HP} and \texttt{A} respectively. It suggests that muscles undergoing very strong degrees of injury must be processed separately, relying \textcolor{black}{either} on pathological data only or manual delineations. Nevetheless, \textcolor{black}{learning scheme} \texttt{A} appears globally better suited from weak to moderately severe muscle impairments.

\begin{figure*}
\hspace{-3cm} \begin{tabular}{ccc}
\hspace{-0.2cm} \includegraphics[height=4.7cm]{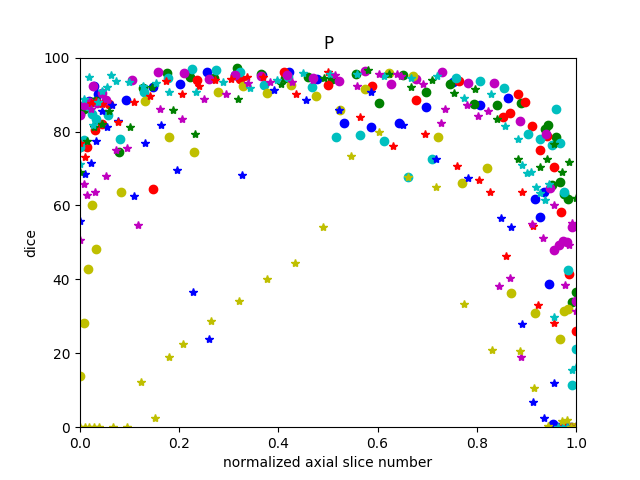} &
\hspace{-0.9cm} \includegraphics[height=4.7cm]{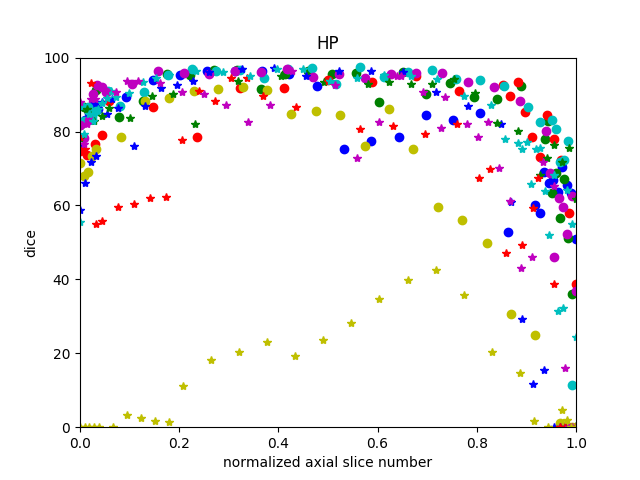} &
\hspace{-0.9cm} \includegraphics[height=4.7cm]{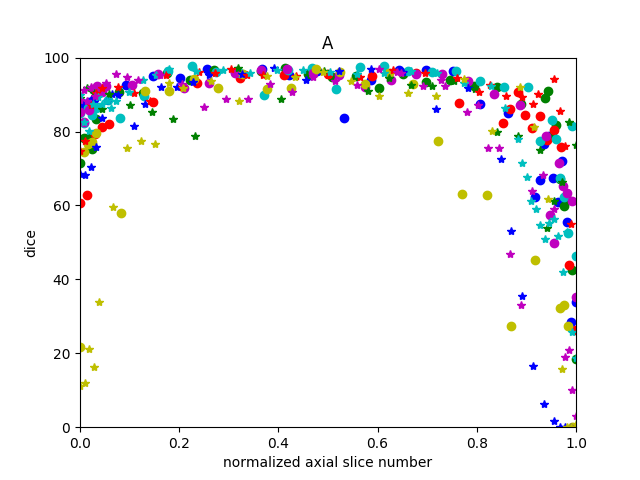} \vspace{-0.2cm} \cr
\hspace{-0.2cm} \includegraphics[height=4.7cm]{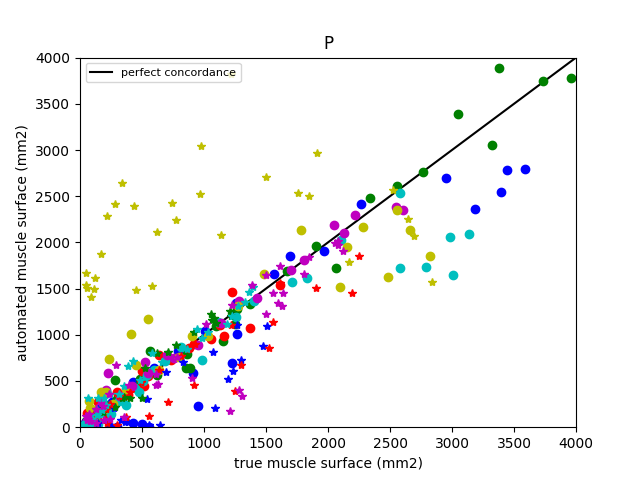} &
\hspace{-0.9cm} \includegraphics[height=4.7cm]{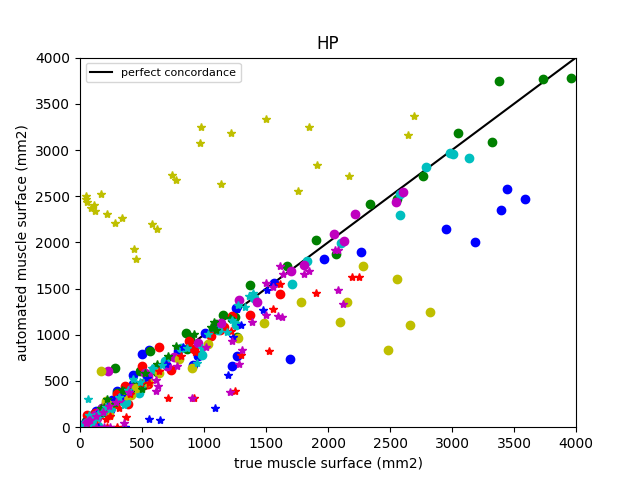} &
\hspace{-0.9cm} \includegraphics[height=4.7cm]{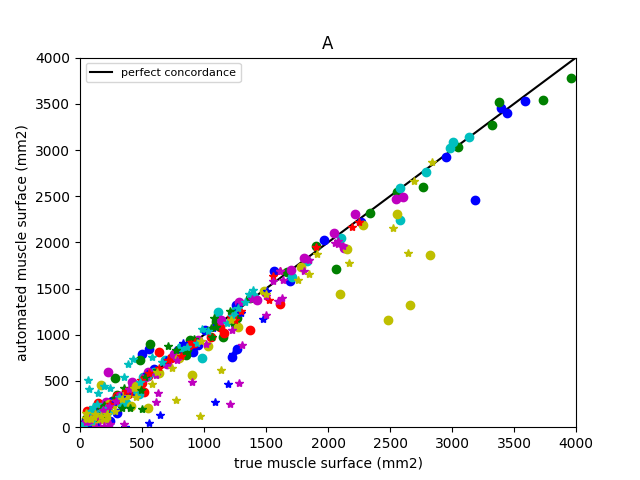} \vspace{0.1cm} \cr
\end{tabular} 
\begin{tabular}{c}
\hspace{-2.4cm} \includegraphics[height=0.20cm]{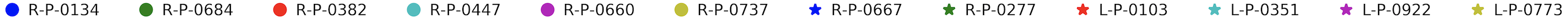} \cr
\end{tabular} 
\caption{Deltoid segmentation accuracy using U-Net \cite{ronneberger2015unet} with learning schemes \texttt{P}, \texttt{HP} and \texttt{A} for each annotated \textcolor{black}{slice} of the whole pathological dataset. Top raw shows Dice scores \textcolor{black}{(\%)} with respect to the normalized axial slice number obtained by linearly scaling slice number from $[z_{min},z_{max}]$ to $[0,1]$ where $\{z_{min},z_{max}$\} are the minimal and maximal axial slice \textcolor{black}{indices} displaying the deltoid. Bottom row displays concordance between groundtruth and predicted deltoid muscle surfaces in mm$^2$. Black line indicates perfect concordance.}
\label{fig::sec4-1-fig-2}
\end{figure*}

Overall, the segmentation results for all three learning schemes are more accurate for mid-muscle regions than for both base and apex, where muscles appear smaller with strong appearance similarities with surrounding tissues (Fig.\ref{fig::sec4-1-fig-2}, top row). Above conclusions (\texttt{A}\hspace{0.05cm}$>$\hspace{0.05cm}\texttt{HP}\hspace{0.05cm}$>$\hspace{0.05cm}\texttt{P}) are confirmed with much more individual Dice \textcolor{black}{scores} grouped on the interval $[75,95\%]$ for \texttt{A}. The concordance between predicted and groundtruth deltoid surfaces (Fig.\ref{fig::sec4-1-fig-2}, bottom row), demonstrates a stronger correlation for \texttt{A} than for \texttt{P} and \texttt{HP} with individual estimations closer to the line of perfect concordance (\texttt{L-P-0773} is the most telling example)\textcolor{black}{, in} agreement with \textcolor{black}{similarity} scores reported for each learning scheme (Tab.\ref{tab::sec4-tab-1}, \textcolor{black}{Fig.\ref{fig::sec4-1-fig-1}}).

\usetikzlibrary{spy,calc}

\begin{figure*}
\hspace{-3cm} \begin{tabular}{cccccccc}

\rotatebox{90}{\small \textcolor{white}{--------------} \texttt{P}} &

\hspace{-0.4cm} \begin{tikzpicture}[spy using outlines={rectangle,yellow,magnification=2,size=0.9cm, connect spies}]
\node {\includegraphics[width=2.3cm]{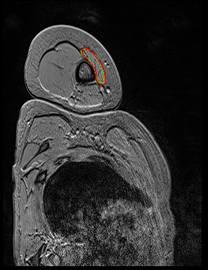}};
\spy on (-0.105,0.762) in node [left] at (1.105,1.005);
\end{tikzpicture} &

\hspace{-0.65cm} \begin{tikzpicture}[spy using outlines={circle,yellow,magnification=3,size=0.9cm, connect spies}]
\node {\includegraphics[width=2.3cm]{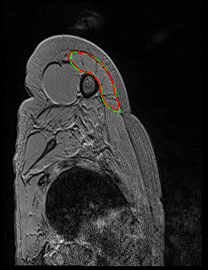}};
\spy on (0.125,0.3) in node [left] at (1.112,1.01);
\end{tikzpicture} &

\hspace{-0.65cm} \begin{tikzpicture}[spy using outlines={circle,yellow,magnification=3,size=0.9cm, connect spies}]
\node {\includegraphics[width=2.3cm]{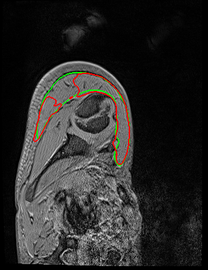}};
\spy on (0.12,-0.08) in node [left] at (1.112,1.01);
\end{tikzpicture} &

\hspace{-0.65cm} \begin{tikzpicture}[spy using outlines={circle,yellow,magnification=3,size=0.9cm, connect spies}]
\node {\includegraphics[width=2.3cm]{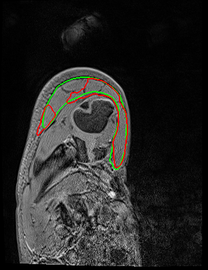}};
\spy on (-0.665,0.09) in node [left] at (1.112,1.01);
\end{tikzpicture} &

\hspace{-0.65cm} \begin{tikzpicture}[spy using outlines={circle,yellow,magnification=3,size=0.9cm, connect spies}]
\node {\includegraphics[width=2.3cm]{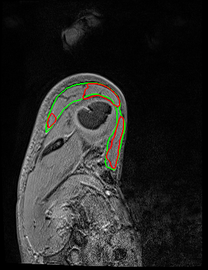}};
\spy on (0.16,0.3) in node [left] at (1.112,1.01);
\end{tikzpicture} &

\hspace{-0.65cm} \begin{tikzpicture}[spy using outlines={circle,yellow,magnification=3,size=0.9cm, connect spies}]
\node {\includegraphics[width=2.3cm]{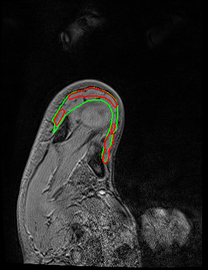}};
\spy on (-0.53,0.09) in node [left] at (1.112,1.01);
\end{tikzpicture} &

\hspace{-0.65cm} \begin{tikzpicture}[spy using outlines={rectangle,yellow,magnification=2,size=0.9cm, connect spies}]
\node {\includegraphics[width=2.3cm]{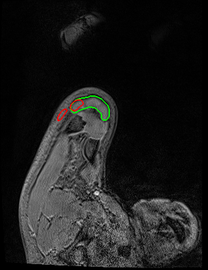}};
\spy on (-0.155,0.31) in node [left] at (1.108,1.008);
\end{tikzpicture} \vspace{-0.25cm} \cr

\rotatebox{90}{\small \textcolor{white}{--------------} \texttt{HP}} &

\hspace{-0.4cm} \begin{tikzpicture}[spy using outlines={rectangle,yellow,magnification=2,size=0.9cm, connect spies}]
\node {\includegraphics[width=2.3cm]{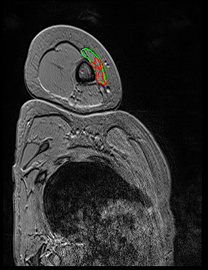}};
\spy on (-0.105,0.762) in node [left] at (1.105,1.005);
\end{tikzpicture} &

\hspace{-0.65cm} \begin{tikzpicture}[spy using outlines={circle,yellow,magnification=3,size=0.9cm, connect spies}]
\node {\includegraphics[width=2.3cm]{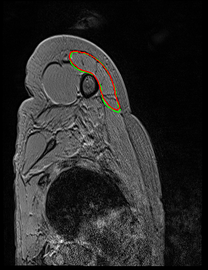}};
\spy on (0.125,0.3) in node [left] at (1.112,1.01);
\end{tikzpicture} &

\hspace{-0.65cm} \begin{tikzpicture}[spy using outlines={circle,yellow,magnification=3,size=0.9cm, connect spies}]
\node {\includegraphics[width=2.3cm]{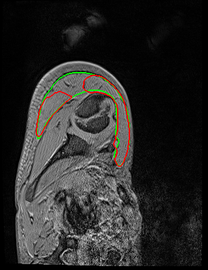}};
\spy on (0.12,-0.08) in node [left] at (1.112,1.01);
\end{tikzpicture} &

\hspace{-0.65cm} \begin{tikzpicture}[spy using outlines={circle,yellow,magnification=3,size=0.9cm, connect spies}]
\node {\includegraphics[width=2.3cm]{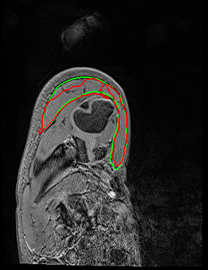}};
\spy on (-0.665,0.09) in node [left] at (1.112,1.01);
\end{tikzpicture} &

\hspace{-0.65cm} \begin{tikzpicture}[spy using outlines={circle,yellow,magnification=3,size=0.9cm, connect spies}]
\node {\includegraphics[width=2.3cm]{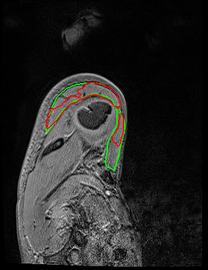}};
\spy on (0.16,0.3) in node [left] at (1.112,1.01);
\end{tikzpicture} &

\hspace{-0.65cm} \begin{tikzpicture}[spy using outlines={circle,yellow,magnification=3,size=0.9cm, connect spies}]
\node {\includegraphics[width=2.3cm]{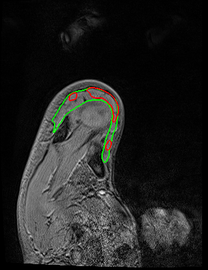}};
\spy on (-0.53,0.09) in node [left] at (1.112,1.01);
\end{tikzpicture} &

\hspace{-0.65cm} \begin{tikzpicture}[spy using outlines={rectangle,yellow,magnification=2,size=0.9cm, connect spies}]
\node {\includegraphics[width=2.3cm]{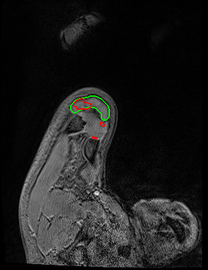}};
\spy on (-0.155,0.31) in node [left] at (1.108,1.008);
\end{tikzpicture} \vspace{-0.25cm} \cr

\rotatebox{90}{\small \textcolor{white}{-------------.} \texttt{A}} &

\hspace{-0.4cm} \begin{tikzpicture}[spy using outlines={rectangle,yellow,magnification=2,size=0.9cm, connect spies}]
\node {\includegraphics[width=2.3cm]{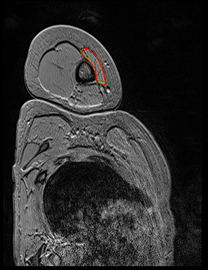}};
\spy on (-0.105,0.762) in node [left] at (1.105,1.005);
\end{tikzpicture} &

\hspace{-0.65cm} \begin{tikzpicture}[spy using outlines={circle,yellow,magnification=3,size=0.9cm, connect spies}]
\node {\includegraphics[width=2.3cm]{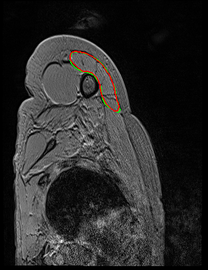}};
\spy on (0.125,0.3) in node [left] at (1.112,1.01);
\end{tikzpicture} &

\hspace{-0.65cm} \begin{tikzpicture}[spy using outlines={circle,yellow,magnification=3,size=0.9cm, connect spies}]
\node {\includegraphics[width=2.3cm]{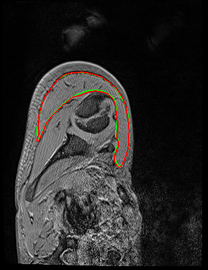}};
\spy on (0.12,-0.08) in node [left] at (1.112,1.01);
\end{tikzpicture} &

\hspace{-0.65cm} \begin{tikzpicture}[spy using outlines={circle,yellow,magnification=3,size=0.9cm, connect spies}]
\node {\includegraphics[width=2.3cm]{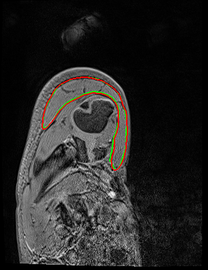}};
\spy on (-0.665,0.09) in node [left] at (1.112,1.01);
\end{tikzpicture} &

\hspace{-0.65cm} \begin{tikzpicture}[spy using outlines={circle,yellow,magnification=3,size=0.9cm, connect spies}]
\node {\includegraphics[width=2.3cm]{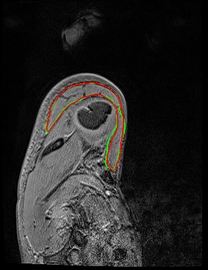}};
\spy on (0.16,0.3) in node [left] at (1.112,1.01);
\end{tikzpicture} &

\hspace{-0.65cm} \begin{tikzpicture}[spy using outlines={circle,yellow,magnification=3,size=0.9cm, connect spies}]
\node {\includegraphics[width=2.3cm]{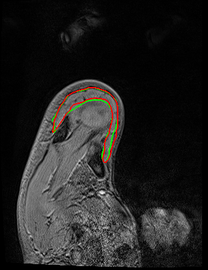}};
\spy on (-0.53,0.09) in node [left] at (1.112,1.01);
\end{tikzpicture} &

\hspace{-0.65cm} \begin{tikzpicture}[spy using outlines={rectangle,yellow,magnification=2,size=0.9cm, connect spies}]
\node {\includegraphics[width=2.3cm]{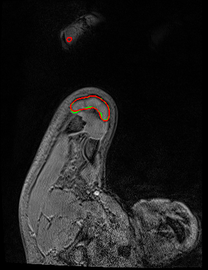}};
\spy on (-0.155,0.31) in node [left] at (1.108,1.008);
\end{tikzpicture} \vspace{-0.23cm} \cr

\end{tabular}
\caption{Automatic pathological deltoid segmentation using U-Net \cite{ronneberger2015unet} embedded with learning schemes \texttt{P}, \texttt{HP} and \texttt{A}. Groundtruth and estimated delineations are in green and red respectively. Displayed results cover the whole muscle spatial extent for \texttt{L-P-0103} examination.}
\label{fig::sec4-1-fig-3}
\end{figure*}

\begin{figure*}
\hspace{-3.25cm} \begin{tabular}{cccccccc}
\rotatebox{90}{\small \textcolor{white}{-----.} infraspinatus} &

\hspace{-0.45cm} \begin{tikzpicture}[spy using outlines={rectangle,yellow,magnification=2,size=1.15cm, connect spies}]
\node {\includegraphics[width=2.4cm]{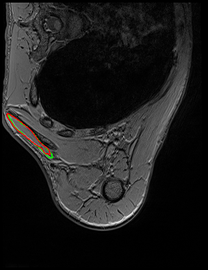}};
\spy on (-0.852,0) in node [left] at (1.158,0.95);
\end{tikzpicture} &

\hspace{-0.65cm} \begin{tikzpicture}[spy using outlines={circle,yellow,magnification=3,size=0.9cm, connect spies}]
\node {\includegraphics[width=2.4cm]{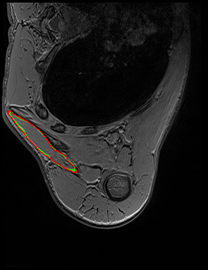}};
\spy on (-0.36,-0.39) in node [left] at (1.158,1.07);
\end{tikzpicture} &

\hspace{-0.65cm} \begin{tikzpicture}[spy using outlines={circle,yellow,magnification=3,size=0.9cm, connect spies}]
\node {\includegraphics[width=2.4cm]{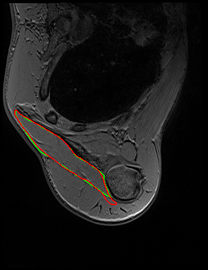}};
\spy on (0.08,-0.73) in node [left] at (1.158,1.07);
\end{tikzpicture} &

\hspace{-0.65cm} \begin{tikzpicture}[spy using outlines={circle,yellow,magnification=3,size=0.9cm, connect spies}]
\node {\includegraphics[width=2.4cm]{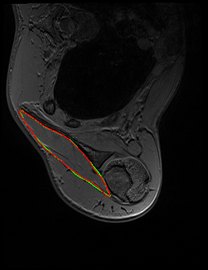}};
\spy on (0.05,-0.65) in node [left] at (1.158,1.07);
\end{tikzpicture} &

\hspace{-0.65cm} \begin{tikzpicture}[spy using outlines={circle,yellow,magnification=3,size=0.9cm, connect spies}]
\node {\includegraphics[width=2.4cm]{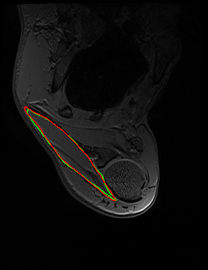}};
\spy on (0.12,-0.72) in node [left] at (1.158,1.07);
\end{tikzpicture} &

\hspace{-0.65cm} \begin{tikzpicture}[spy using outlines={circle,yellow,magnification=3,size=0.9cm, connect spies}]
\node {\includegraphics[width=2.4cm]{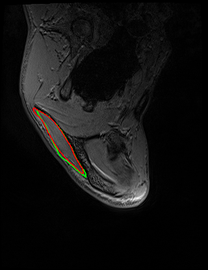}};
\spy on (-0.24,-0.43) in node [left] at (1.158,1.07);
\end{tikzpicture} &

\hspace{-0.65cm} \begin{tikzpicture}[spy using outlines={rectangle,yellow,magnification=3,size=1cm, connect spies}]
\node {\includegraphics[width=2.4cm]{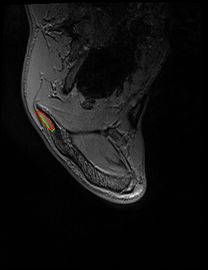}};
\spy on (-0.67,0.165) in node [left] at (1.158,1.02);
\end{tikzpicture} \vspace{-0.25cm} \cr

\rotatebox{90}{\small \textcolor{white}{-----.} supraspinatus} &

\hspace{-0.45cm} \begin{tikzpicture}[spy using outlines={rectangle,yellow,magnification=2,size=1cm, connect spies}]
\node {\includegraphics[width=2.4cm]{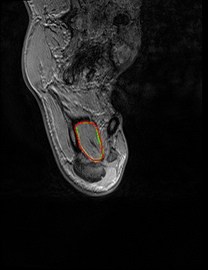}};
\spy on (-0.17,-0.078) in node [left] at (1.158,-1.02);
\end{tikzpicture} &

\hspace{-0.65cm} \begin{tikzpicture}[spy using outlines={rectangle,yellow,magnification=2,size=1.15cm, connect spies}]
\node {\includegraphics[width=2.4cm]{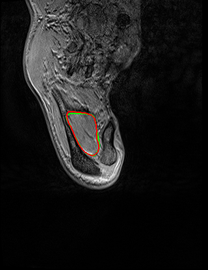}};
\spy on (-0.25,0.015) in node [left] at (1.158,-0.95);
\end{tikzpicture} &

\hspace{-0.65cm} \begin{tikzpicture}[spy using outlines={rectangle,yellow,magnification=2,size=1.22cm, connect spies}]
\node {\includegraphics[width=2.4cm]{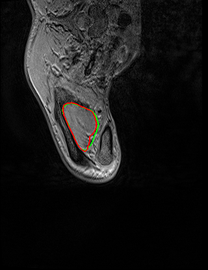}};
\spy on (-0.27,0.09) in node [left] at (1.158,-0.91);
\end{tikzpicture} &

\hspace{-0.65cm} \begin{tikzpicture}[spy using outlines={rectangle,yellow,magnification=2,size=1.15cm, connect spies}]
\node {\includegraphics[width=2.4cm]{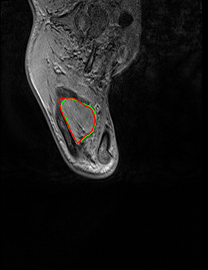}};
\spy on (-0.3,0.17) in node [left] at (1.158,-0.95);
\end{tikzpicture} &

\hspace{-0.65cm} \begin{tikzpicture}[spy using outlines={rectangle,yellow,magnification=2,size=1.15cm, connect spies}]
\node {\includegraphics[width=2.4cm]{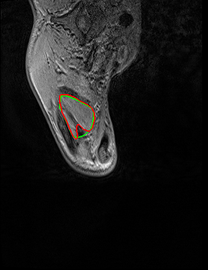}};
\spy on (-0.3,0.21) in node [left] at (1.158,-0.95);
\end{tikzpicture} &

\hspace{-0.65cm} \begin{tikzpicture}[spy using outlines={rectangle,yellow,magnification=2,size=1cm, connect spies}]
\node {\includegraphics[width=2.4cm]{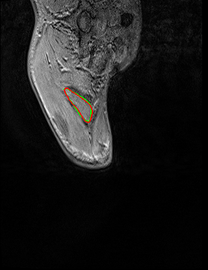}};
\spy on (-0.28,0.34) in node [left] at (1.158,-1.02);
\end{tikzpicture} &

\hspace{-0.65cm} \begin{tikzpicture}[spy using outlines={rectangle,yellow,magnification=2,size=1cm, connect spies}]
\node {\includegraphics[width=2.4cm]{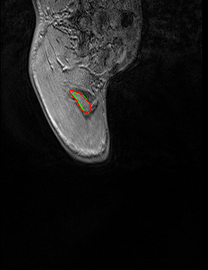}};
\spy on (-0.25,0.38) in node [left] at (1.158,-1.02);
\end{tikzpicture} \vspace{-0.25cm} \cr

\rotatebox{90}{\small \textcolor{white}{------} subscapularis} &

\hspace{-0.45cm} \begin{tikzpicture}[spy using outlines={circle,yellow,magnification=3,size=0.9cm, connect spies}]
\node {\includegraphics[width=2.4cm]{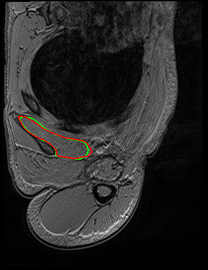}};
\spy on (-0.95,0.2) in node [left] at (1.158,1.07);
\end{tikzpicture} &

\hspace{-0.65cm} \begin{tikzpicture}[spy using outlines={circle,yellow,magnification=3,size=0.9cm, connect spies}]
\node {\includegraphics[width=2.4cm]{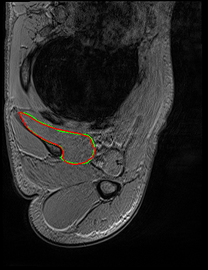}};
\spy on (-0.94,0.23) in node [left] at (1.158,1.07);
\end{tikzpicture} &

\hspace{-0.65cm} \begin{tikzpicture}[spy using outlines={circle,yellow,magnification=3,size=0.9cm, connect spies}]
\node {\includegraphics[width=2.4cm]{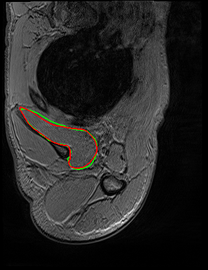}};
\spy on (-0.93,0.29) in node [left] at (1.158,1.07);
\end{tikzpicture} &

\hspace{-0.65cm} \begin{tikzpicture}[spy using outlines={circle,yellow,magnification=3,size=0.9cm, connect spies}]
\node {\includegraphics[width=2.4cm]{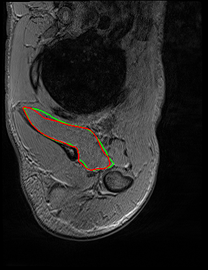}};
\spy on (-0.88,0.3) in node [left] at (1.158,1.07);
\end{tikzpicture} &

\hspace{-0.65cm} \begin{tikzpicture}[spy using outlines={circle,yellow,magnification=3,size=0.9cm, connect spies}]
\node {\includegraphics[width=2.4cm]{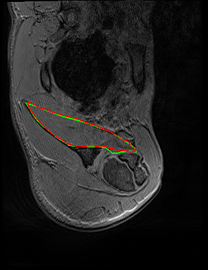}};
\spy on (-0.825,0.34) in node [left] at (1.158,1.07);
\end{tikzpicture} &

\hspace{-0.65cm} \begin{tikzpicture}[spy using outlines={circle,yellow,magnification=3,size=0.9cm, connect spies}]
\node {\includegraphics[width=2.4cm]{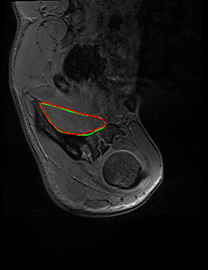}};
\spy on (-0.7,0.35) in node [left] at (1.158,1.07);
\end{tikzpicture} &

\hspace{-0.65cm} \begin{tikzpicture}[spy using outlines={rectangle,yellow,magnification=2,size=1cm, connect spies}]
\node {\includegraphics[width=2.4cm]{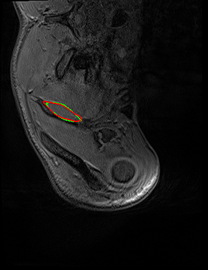}};
\spy on (-0.48,0.28) in node [left] at (1.162,1.02);
\end{tikzpicture}  \vspace{-0.15cm} \cr

\end{tabular}
\caption{Automatic pathological segmentation of infraspinatus, supraspinatus and subscapularis using U-Net \cite{ronneberger2015unet} with training on both healthy and pathological data simultaneously (\texttt{A}). Groundtruth and estimated delineations are in green and red respectively. Displayed results cover the whole muscle spatial extents for \texttt{R-P-0447} \textcolor{black}{(top)}, \texttt{R-P-0660} \textcolor{black}{(middle)} and \texttt{R-P-0134} \textcolor{black}{(bottom)} examinations.}
\label{fig::sec4-1-fig-4}
\end{figure*}

Visually comparing both manual and automatic segmentation for deltoid (\texttt{P}, \texttt{HP} and \texttt{A}, Fig.\ref{fig::sec4-1-fig-3}) and other rotator cuff muscles (\texttt{A} only, Fig.\ref{fig::sec4-1-fig-4}) further supports the validity of \textcolor{black}{automatic} segmentation. A very accurate deltoid delineation is achieved for \texttt{A} whereas \texttt{P} and \texttt{HP} tend to under-segment the muscle area (Fig.\ref{fig::sec4-1-fig-3}). Complex muscle shapes and subtle contours (Fig.\ref{fig::sec4-1-fig-4}) are relatively well captured. In addition, we can notice outstanding performance near muscle insertion regions (Fig.\ref{fig::sec4-1-fig-4}) whose contours are usually very hard to extract\textcolor{black}{, even visually}. These results confirm that using simultaneously healthy and pathological data for training helps in providing good model generalizability despite the data scarcity issue combined with a large appearance variability.

\subsection{Extended architectures with pre-trained encoders}
\label{sec:sec4-2}

The \texttt{v16pU-Net} architecture globally outperforms \textcolor{black}{both} U-Net and \texttt{v16U-Net} \textcolor{black}{networks (Fig.\ref{fig::sec4-1-fig-1})} with Dice scores of $82.42\%$ for deltoid, $81.98\%$ for infraspinatus, $70.98\%$ for supraspinatus and 82.80\% for subscapularis (Tab.\ref{tab::sec4-tab-1}). On the contrary, \texttt{v16U-Net} (\textcolor{black}{U-Net}) obtains $80.05\%$ ($78.32\%$) for deltoid, $81.91\%$ ($81.58\%$) for infraspinatus, $67.30\%$ ($65.68\%$) for supraspinatus and $81.58$\% ($81.41$\%) for subscapularis. In one hand, despite slightly worse scores compared with U-Net for infraspinatus in terms of sensitivity ($83.74$ against $84.61\textcolor{black}{\%}$) and \texttt{ASE} ($80.11$ against $74.47$mm$^2$), \texttt{v16U-Net} is most likely to provide good predictive performance and model generalizability thanks to its deeper architecture. On the other hand, comparisons between \texttt{v16U-Net} and \texttt{v16pU-Net} \textcolor{black}{reveal} that pre-training the encoder using ImageNet brings non-negligible improvements \textcolor{black}{(Fig.\ref{fig::sec4-1-fig-1})}. For instance, \texttt{v16pU-Net} provides significant gains \textcolor{black}{(Tab.\ref{tab::sec4-tab-1})} for deltoid (\textcolor{black}{supraspinatus}) whose Jaccard score goes from $71.46$ ($56.98$) to $74\%$ ($61.31\%$). \textcolor{black}{The Cohen's kappa coefficient enhancement is around $2.4\%$ ($3.7\%$).} Surface estimation errors are among the lowest obtained with only $80.38$mm$^2$ for deltoid and 82.95mm$^2$ for subscapularis. \textcolor{black}{Medians and first quartiles (Fig.\ref{fig::sec4-1-fig-1}) globally highlight significant segmentation gains, especially for supraspinatus}. Despite their non-medical nature, the large amount of ImageNet images used for pre-training makes the network converge towards a better solution. \texttt{v16pU-Net} is therefore the most able to efficiently discriminate individual muscles from surrounding anatomical structures, compared to U-Net and \texttt{v16U-Net}. In average among the four shoulder muscles, gains for Dice, \textcolor{black}{sensitivity,} Jaccard \textcolor{black}{and kappa} reach $2.8$, $2.7$\textcolor{black}{, $3.2$ and $2.8\%$} from U-Net to \texttt{v16pU-Net}. 

\begin{table*}
\small 
\centering \begin{tabular}{|c|l|c|c|c?c|}
\hline
\multirow{2}{*}{metric} & \multicolumn{1}{c|}{scheme} & \texttt{P} & \texttt{HP} & \multicolumn{2}{c|}{\texttt{A}} \\ \cline{2-6} 
                        & \multicolumn{1}{c|}{network} & \multicolumn{3}{c?}{\texttt{U-Net} \cite{ronneberger2015unet}} & \texttt{v16U-Net} \\ \hline \hline
\multirow{4}{*}{\rotatebox{90}{\texttt{dice}}} & deltoid & \textbf{3.7}$\times10^{-29}$ & \textbf{8.8}$\times10^{-21}$ & \textbf{9.7}$\times10^{-11}$ & \textbf{7.5}$\times10^{-6}$ \\ \cline{2-6} 
& infraspinatus & \textbf{2.3}$\times10^{-19}$ & \textbf{5.4}$\times10^{-7}$ & 0.491 & 0.935 \\ \cline{2-6} 
& supraspinatus & \textbf{4.3}$\times10^{-6}$ & \textbf{4.8}$\times10^{-5}$ & \textbf{3.2}$\times10^{-6}$ & \textbf{7.0}$\times10^{-5}$ \\ \cline{2-6} 
& subscapularis & \textbf{1.1}$\times10^{-13}$ & \textbf{1.5}$\times10^{-7}$ & \textbf{0.001} & \textbf{0.006} \\ \hline \hline
\multirow{4}{*}{\rotatebox{90}{\texttt{sens}}} & deltoid & \textbf{1.2}$\times10^{-23}$ & \textbf{3.7}$\times10^{-23}$ & \textbf{2.2}$\times10^{-9}$ & \textbf{4.4}$\times10^{-4}$ \\ \cline{2-6} 
& infraspinatus & \textbf{6.1}$\times10^{-16}$ & \textbf{3.9}$\times10^{-4}$ & 0.117 & 0.788 \\ \cline{2-6} 
& supraspinatus & \textbf{8.6}$\times10^{-4}$ & \textbf{8.7}$\times10^{-5}$ & \textbf{0.016} & 0.135 \\ \cline{2-6} 
& subscapularis & \textbf{2.7}$\times10^{-12}$ & \textbf{2.8}$\times10^{-10}$ & \textbf{0.002} & \textbf{1.5}$\times10^{-6}$ \\ \hline \hline
\multirow{4}{*}{\rotatebox{90}{\texttt{spec}}} & deltoid & \textbf{7.9}$\times10^{-9}$ & \textbf{2.6}$\times10^{-7}$ & 0.457 & 0.069 \\ \cline{2-6} 
& infraspinatus & \textbf{1.3}$\times10^{-5}$ & \textbf{1.2}$\times10^{-6}$ & \textbf{0.010} & 0.387 \\ \cline{2-6} 
& supraspinatus & \textbf{8.2}$\times10^{-6}$ & 0.118 & \textbf{1.0}$\times10^{-5}$ & \textbf{7.8}$\times10^{-9}$ \\ \cline{2-6} 
& subscapularis & 0.924 & \textbf{0.005} & 0.078 & \textbf{2.5}$\times10^{-4}$ \\ \hline \hline
\multirow{4}{*}{\rotatebox{90}{\texttt{jacc}}} & deltoid & \textbf{5.1}$\times10^{-35}$ & \textbf{9.9}$\times10^{-25}$ & \textbf{7.0}$\times10^{-11}$ & \textbf{1.1}$\times10^{-5}$ \\ \cline{2-6} 
& infraspinatus & \textbf{1.6}$\times10^{-23}$ & \textbf{1.6}$\times10^{-9}$ & 0.251 & 0.917 \\ \cline{2-6} 
& supraspinatus & \textbf{4.9}$\times10^{-10}$ & \textbf{8.3}$\times10^{-7}$ & \textbf{1.8}$\times10^{-7}$ & \textbf{1.7}$\times10^{-6}$ \\ \cline{2-6} 
& subscapularis & \textbf{1.2}$\times10^{-17}$ & \textbf{1.0}$\times10^{-9}$ & \textbf{2.6}$\times10^{-4}$ & \textbf{0.002} \\ \hline \hline
\multirow{4}{*}{\rotatebox{90}{\texttt{kappa}}} & deltoid & \textbf{2.6}$\times10^{-29}$ & \textbf{7.9}$\times10^{-21}$ & \textbf{8.9}$\times10^{-11}$ & \textbf{6.8}$\times10^{-6}$ \\ \cline{2-6} 
& infraspinatus & \textbf{1.7}$\times10^{-19}$ & \textbf{4.6}$\times10^{-7}$ & 0.476 & 0.928 \\ \cline{2-6} 
& supraspinatus & \textbf{3.4}$\times10^{-6}$ & \textbf{4.3}$\times10^{-5}$ & \textbf{2.9}$\times10^{-6}$ & \textbf{5.8}$\times10^{-5}$ \\ \cline{2-6} 
& subscapularis & \textbf{9.5}$\times10^{-14}$ & \textbf{1.4}$\times10^{-7}$ & \textbf{0.001} & \textbf{0.006} \\ \hline \hline
\multirow{4}{*}{\rotatebox{90}{\texttt{ASE}}} & deltoid & \textbf{2.2}$\times10^{-15}$ & \textbf{7.9}$\times10^{-13}$ & \textbf{0.001} & \textbf{0.021} \\ \cline{2-6} 
& infraspinatus & \textbf{8.4}$\times10^{-10}$ & \textbf{0.021} & 0.308 & 0.813 \\ \cline{2-6} 
& supraspinatus & \textbf{7.3}$\times10^{-5}$ & \textbf{8.2}$\times10^{-4}$ & \textbf{0.004} & \textbf{0.036} \\ \cline{2-6} 
& subscapularis & \textbf{0.009} & \textbf{0.009} & \textbf{0.005} & \textbf{0.011} \\ \hline
\end{tabular}
\caption{\textcolor{black}{Statistical analysis between \texttt{v16pU-Net} embedded with learning scheme \texttt{A} and all other configurations (U-Net \cite{ronneberger2015unet} with \texttt{P}, \texttt{HP} and \texttt{A} as well as \texttt{v16U-Net} with \texttt{A}) through Student’s paired t-tests using Dice, sensitivity, specificity, Jaccard, Cohen's kappa scores as well as absolute surface error over the pathological dataset. Bold p-values ($<0.05$) highlight statistically significant results.}}
\label{tab::sec4-tab-2}
\end{table*}

\textcolor{black}{Above conclusions (\texttt{v16pU-Net}\hspace{0.05cm}$>$\hspace{0.05cm}\texttt{v16U-Net}\hspace{0.05cm}$>$\hspace{0.05cm}U-Net) are further supported by statistical analysis (Tab.\ref{tab::sec4-tab-2}). Except for infraspinatus, Student’s paired t-tests between \texttt{v16pU-Net} and \texttt{v16U-Net} or U-Net globally indicate that extended architectures with pre-trained encoders really bring non-negligible improvements (p-values $<0.05$ for similarity metrics and \texttt{ASE}). This finding is all the more verified between \texttt{v16pU-Net} embedded with learning scheme \texttt{A} and U-Net \cite{ronneberger2015unet} with \texttt{P}, \texttt{HP} or \texttt{A} for all muscles including infraspinatus.}

\begin{figure*}
\hspace{-3cm} \begin{tabular}{ccc}
\hspace{-0.2cm} \includegraphics[height=4.7cm]{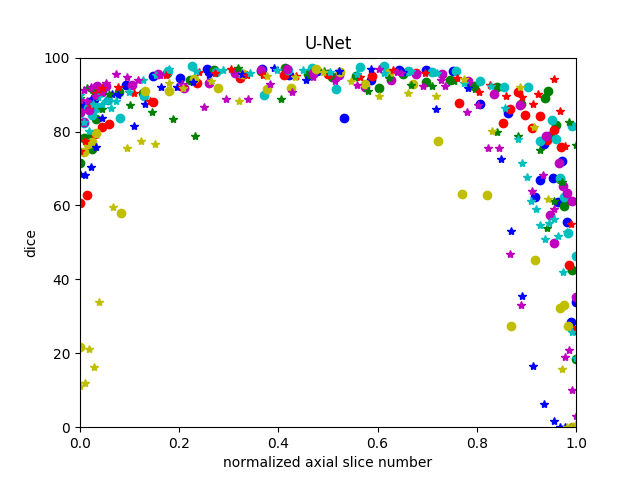} &
\hspace{-0.9cm} \includegraphics[height=4.7cm]{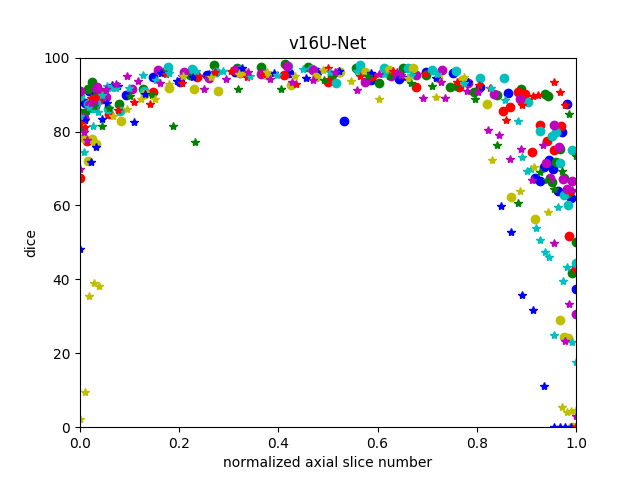} &
\hspace{-0.9cm} \includegraphics[height=4.7cm]{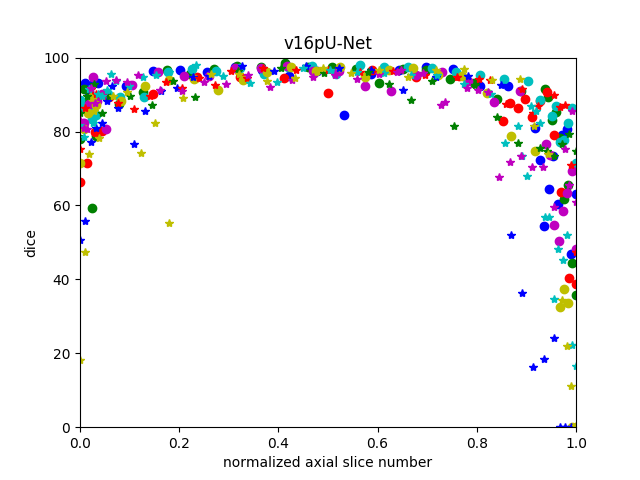} \vspace{-0.2cm} \cr
\hspace{-0.2cm} \includegraphics[height=4.7cm]{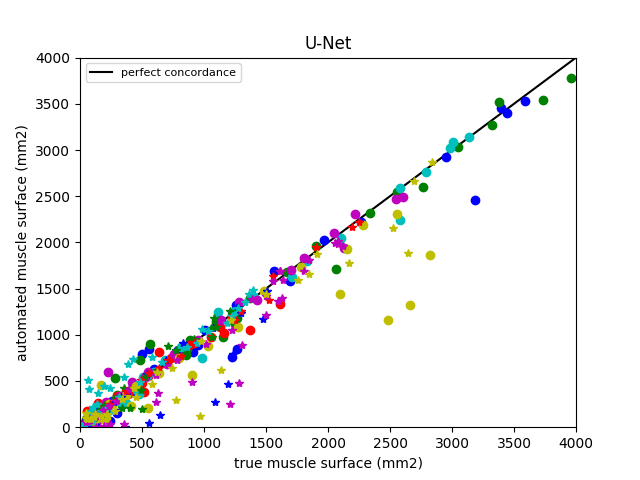} &
\hspace{-0.9cm} \includegraphics[height=4.7cm]{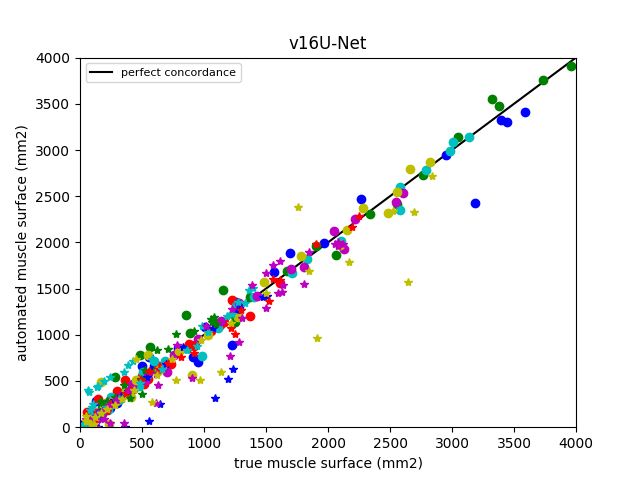} &
\hspace{-0.9cm} \includegraphics[height=4.7cm]{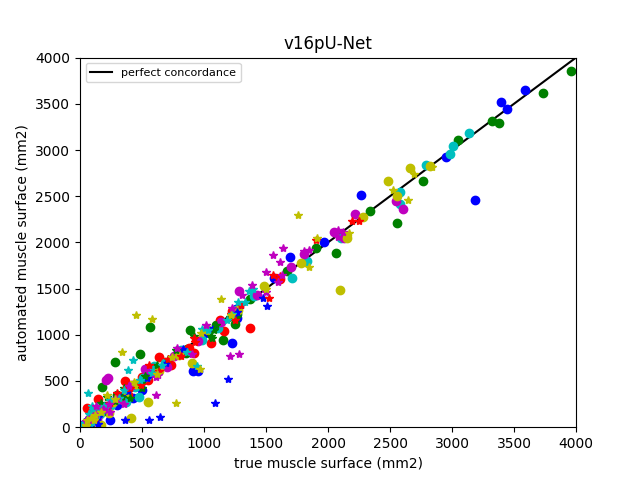} \vspace{0.1cm} \cr
\end{tabular} 
\begin{tabular}{c}
\hspace{-2.4cm}  \includegraphics[height=0.20cm]{./figures/legend-examinations.png} \cr
\end{tabular} 
\caption{Deltoid segmentation accuracy using U-Net \cite{ronneberger2015unet}, \texttt{v16U-Net} and \texttt{v16pU-Net} with learning scheme \texttt{A} for each annotated \textcolor{black}{slice} of the whole pathological dataset. Top raw shows Dice scores with respect to normalized axial slice number. Bottom row displays concordance between groundtruth and predicted deltoid surfaces. Black line indicates perfect concordance.}
\label{fig::sec4-2-fig-1}
\end{figure*}

From U-Net to \texttt{v16pU-Net}, individual Dice scores (Fig.\ref{fig::sec4-2-fig-1}, top row) are slightly pushed towards the upper limit ($100\%$) with less variability and an increased overall consistency along \textcolor{black}{the axial axis}, as for \texttt{R-P-0737} and \texttt{L-P-0773}. Extreme axial slices are much better handled in the case \texttt{v16pU-Net}, especially when normalized slice numbers approach zero. In addition, a slightly stronger correlation between predicted and groundtruth deltoid surface can be seen for \texttt{v16pU-Net}  with respect to U-Net and \texttt{v16U-Net} (Fig.\ref{fig::sec4-2-fig-1}, bottom row). In particular, great improvements for \texttt{R-P-0737} and \texttt{L-P-0773} can be highlighted. 

\begin{figure*}
\hspace{-3.25cm} \begin{tabular}{ccccccccc}
& \multicolumn{2}{c}{deltoid} & \multicolumn{2}{c}{infraspinatus} & \multicolumn{2}{c}{supraspinatus} & \multicolumn{2}{c}{subscapularis} \vspace{0.05cm} \cr

\rotatebox{90}{\small \textcolor{white}{------} U-Net \cite{ronneberger2015unet}} & 

\hspace{-0.45cm} \begin{tikzpicture}[spy using outlines={circle,yellow,magnification=3,size=0.9cm, connect spies}]
\node {\includegraphics[width=2.08cm]{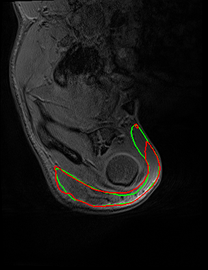}};
\spy on (0.3,0.06) in node [left] at (1,0.86);
\end{tikzpicture} &

\hspace{-0.65cm} \begin{tikzpicture}[spy using outlines={circle,yellow,magnification=3,size=0.9cm, connect spies}]
\node {\includegraphics[width=2.08cm]{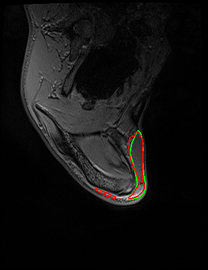}};
\spy on (0.35,-0.02) in node [left] at (1,0.86);
\end{tikzpicture} &

\hspace{-0.65cm} \begin{tikzpicture}[spy using outlines={rectangle,yellow,magnification=2,size=0.9cm, connect spies}]
\node {\includegraphics[width=2.08cm]{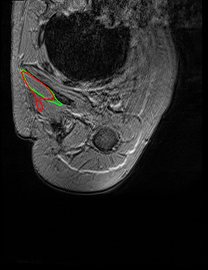}};
\spy on (-0.62,0.45) in node [left] at (1,-0.86);
\end{tikzpicture} &

\hspace{-0.65cm} \begin{tikzpicture}[spy using outlines={circle,yellow,magnification=2,size=0.9cm, connect spies}]
\node {\includegraphics[width=2.08cm]{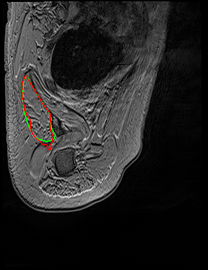}};
\spy on (-0.55,-0.02) in node [left] at (1,-0.86);
\end{tikzpicture} &

\hspace{-0.65cm} \begin{tikzpicture}[spy using outlines={rectangle,yellow,magnification=2,size=0.9cm, connect spies}]
\node { \includegraphics[width=2.08cm]{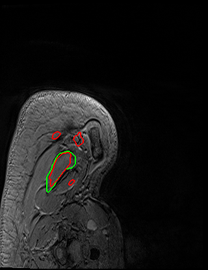}};
\spy on (-0.42,-0.365) in node [left] at (1,0.86);
\end{tikzpicture} &

\hspace{-0.65cm} \begin{tikzpicture}[spy using outlines={circle,yellow,magnification=3,size=0.9cm, connect spies}]
\node {\includegraphics[width=2.08cm]{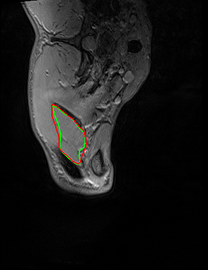}};
\spy on (-0.23,-0.23) in node [left] at (1,-0.86);
\end{tikzpicture} &

\hspace{-0.65cm} \begin{tikzpicture}[spy using outlines={circle,yellow,magnification=3,size=0.9cm, connect spies}]
\node {\includegraphics[width=2.08cm]{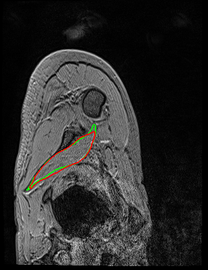}};
\spy on (-0.1,0.05) in node [left] at (1,0.86);
\end{tikzpicture} &

\hspace{-0.65cm} \begin{tikzpicture}[spy using outlines={circle,yellow,magnification=3,size=0.9cm, connect spies}]
\node {\includegraphics[width=2.08cm]{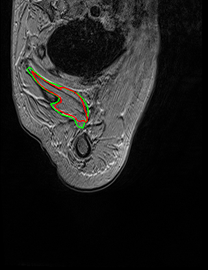}};
\spy on (-0.2,0.15) in node [left] at (1,-0.86);
\end{tikzpicture} \vspace{-0.2cm} \cr 

\rotatebox{90}{\small \textcolor{white}{------} \texttt{v16U-Net}} & 

\hspace{-0.45cm} \begin{tikzpicture}[spy using outlines={circle,yellow,magnification=3,size=0.9cm, connect spies}]
\node {\includegraphics[width=2.08cm]{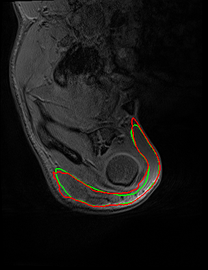}};
\spy on (0.3,0.06) in node [left] at (1,0.86);
\end{tikzpicture} &

\hspace{-0.65cm} \begin{tikzpicture}[spy using outlines={circle,yellow,magnification=3,size=0.9cm, connect spies}]
\node {\includegraphics[width=2.08cm]{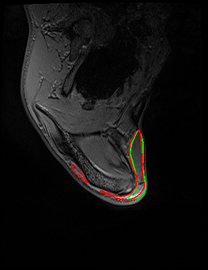}};
\spy on (0.35,-0.02) in node [left] at (1,0.86);
\end{tikzpicture} &

\hspace{-0.65cm} \begin{tikzpicture}[spy using outlines={rectangle,yellow,magnification=2,size=0.9cm, connect spies}]
\node {\includegraphics[width=2.08cm]{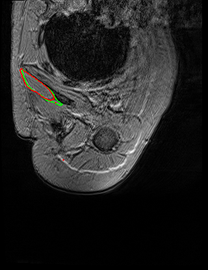}};
\spy on (-0.62,0.45) in node [left] at (1,-0.86);
\end{tikzpicture} &

\hspace{-0.65cm} \begin{tikzpicture}[spy using outlines={circle,yellow,magnification=2,size=0.9cm, connect spies}]
\node {\includegraphics[width=2.08cm]{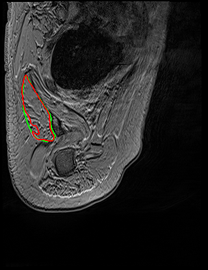}};
\spy on (-0.55,-0.02) in node [left] at (1,-0.86);
\end{tikzpicture} &

\hspace{-0.65cm} \begin{tikzpicture}[spy using outlines={rectangle,yellow,magnification=2,size=0.9cm, connect spies}]
\node {\includegraphics[width=2.08cm]{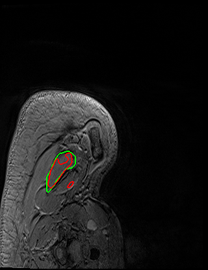}};
\spy on (-0.42,-0.365) in node [left] at (1,0.86);
\end{tikzpicture} &

\hspace{-0.65cm} \begin{tikzpicture}[spy using outlines={circle,yellow,magnification=3,size=0.9cm, connect spies}]
\node {\includegraphics[width=2.08cm]{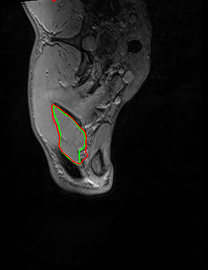}};
\spy on (-0.23,-0.23) in node [left] at (1,-0.86);
\end{tikzpicture} &

\hspace{-0.65cm} \begin{tikzpicture}[spy using outlines={circle,yellow,magnification=3,size=0.9cm, connect spies}]
\node {\includegraphics[width=2.08cm]{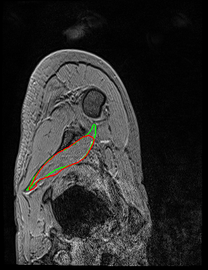}};
\spy on (-0.1,0.05) in node [left] at (1,0.86);
\end{tikzpicture} &

\hspace{-0.65cm} \begin{tikzpicture}[spy using outlines={circle,yellow,magnification=3,size=0.9cm, connect spies}]
\node {\includegraphics[width=2.08cm]{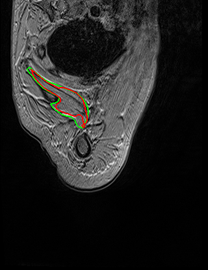}};
\spy on (-0.2,0.15) in node [left] at (1,-0.86);
\end{tikzpicture} \vspace{-0.2cm} \cr 

\rotatebox{90}{\small \textcolor{white}{-----} \texttt{v16pU-Net}} & 

\hspace{-0.45cm} \begin{tikzpicture}[spy using outlines={circle,yellow,magnification=3,size=0.9cm, connect spies}]
\node {\includegraphics[width=2.08cm]{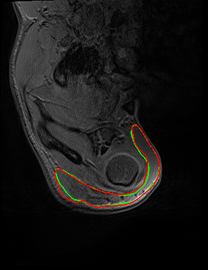}};
\spy on (0.3,0.06) in node [left] at (1,0.86);
\end{tikzpicture} &

\hspace{-0.65cm} \begin{tikzpicture}[spy using outlines={circle,yellow,magnification=3,size=0.9cm, connect spies}]
\node {\includegraphics[width=2.08cm]{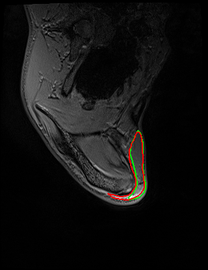}};
\spy on (0.35,-0.02) in node [left] at (1,0.86);
\end{tikzpicture} &

\hspace{-0.65cm} \begin{tikzpicture}[spy using outlines={rectangle,yellow,magnification=2,size=0.9cm, connect spies}]
\node {\includegraphics[width=2.08cm]{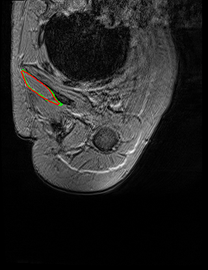}};
\spy on (-0.62,0.45) in node [left] at (1,-0.86);
\end{tikzpicture} &

\hspace{-0.65cm} \begin{tikzpicture}[spy using outlines={circle,yellow,magnification=2,size=0.9cm, connect spies}]
\node {\includegraphics[width=2.08cm]{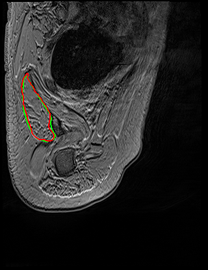}};
\spy on (-0.55,-0.02) in node [left] at (1,-0.86);
\end{tikzpicture} &

\hspace{-0.65cm} \begin{tikzpicture}[spy using outlines={rectangle,yellow,magnification=2,size=0.9cm, connect spies}]
\node {\includegraphics[width=2.08cm]{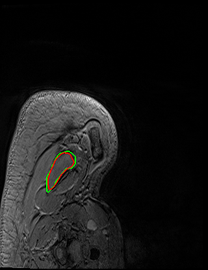}};
\spy on (-0.42,-0.365) in node [left] at (1,0.86);
\end{tikzpicture} &

\hspace{-0.65cm} \begin{tikzpicture}[spy using outlines={circle,yellow,magnification=3,size=0.9cm, connect spies}]
\node {\includegraphics[width=2.08cm]{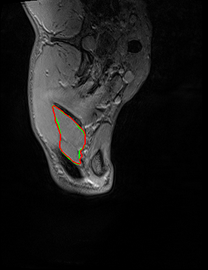}};
\spy on (-0.23,-0.23) in node [left] at (1,-0.86);
\end{tikzpicture} &

\hspace{-0.65cm} \begin{tikzpicture}[spy using outlines={circle,yellow,magnification=3,size=0.9cm, connect spies}]
\node {\includegraphics[width=2.08cm]{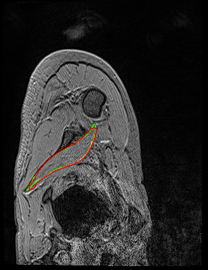}};
\spy on (-0.1,0.05) in node [left] at (1,0.86);
\end{tikzpicture} &

\hspace{-0.65cm} \begin{tikzpicture}[spy using outlines={circle,yellow,magnification=3,size=0.9cm, connect spies}]
\node {\includegraphics[width=2.08cm]{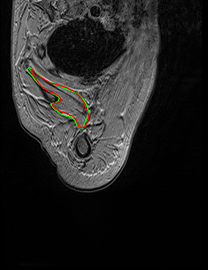}};
\spy on (-0.2,0.15) in node [left] at (1,-0.86);
\end{tikzpicture} \vspace{-0.225cm} \cr 
&
\hspace{-0.43cm} \scriptsize \texttt{R-P-0134} &
\hspace{-0.43cm} \scriptsize \texttt{R-P-0447} &
\hspace{-0.43cm} \scriptsize \texttt{R-P-0667} &
\hspace{-0.43cm} \scriptsize \texttt{R-P-0737} &
\hspace{-0.43cm} \scriptsize \texttt{L-P-0773} &
\hspace{-0.43cm} \scriptsize \texttt{R-P-0684} &
\hspace{-0.43cm} \scriptsize \texttt{L-P-0103} &
\hspace{-0.43cm} \scriptsize \texttt{R-P-0277} \cr

\end{tabular}
\caption{Automatic pathological segmentation of deltoid, infraspinatus, supraspinatus and subscapularis using U-Net \cite{ronneberger2015unet}, \texttt{v16U-Net} and \texttt{v16pU-Net} with training on both healthy and pathological data simultaneously (\texttt{A}). Groundtruth and estimated delineations are in green and red respectively. $8$ pathological examinations among the $12$ available are involved to \textcolor{black}{provide} \textcolor{black}{valuable insight into the overall performance}.} \vspace{-0.3cm} 
\label{fig::sec4-2-fig-2}
\end{figure*}

Globally, compared to U-Net and \texttt{v16U-Net}, better contour adherence and shape consistency are reached by \texttt{v16pU-Net} whose ability to mimmic expert annotations is notable (Fig.\ref{fig::sec4-2-fig-2}). The great diversity in terms of textures (smooth in \texttt{R-P-0684} versus granular in \texttt{R-P-0737}) is accurately captured despite high similar visual properties with surrounding structures. Visual results also reveal that \texttt{v16pU-Net} has a good behavior for complex muscle insertion regions (\texttt{R-P-0447}). Despite a satisfactory overall quality, U-Net and \texttt{v16U-Net} are frequently prone to under- (\texttt{R-P-0134}, \texttt{R-P-0277}) or over-segmentation (\texttt{R-P-0684}). Some examples report inconsistent shapes (\texttt{R-P-0667}, \texttt{R-P-0737}), sometimes combined with false positive areas which can be located far away from the groundtruth muscle location (\texttt{R-P-0447}, \texttt{L-P-0773}). Using a pre-trained and complex architecture such as \texttt{v16pU-Net} to \textcolor{black}{simultaneously} process healthy and pathological data provides accurate \textcolor{black}{automated} delineations of pathological shoulder muscles for patients with OPBB.

\subsection{\textcolor{black}{Benefits for clinical practice}}
\label{sec:sec4-3}

\textcolor{black}{The key contribution of this work deals with the possibility of automatically providing robust MR delineations for shoulder pathological muscles, despite the strong diversity in shape, size, location, texture and injury (Fig.\ref{fig::sec4-2-fig-2}). First, it has the advantages of reducing the burden of manual segmentation and avoiding the subjectivity of experts. Second, it paves the way for the automated inference of individual morphological parameters \cite{pons2017shoulder} which are not accessible with simple clinical examinations. This can therefore be useful to guide the rehabilitative and surgical management of children with OBPP. The benefit of the proposed technology in real clinical use can be also involved for other very frequent shoulder muscular disorders such as rotator cuff tears in order to provide objective predictors of successful surgical repair \cite{laron2012muscle}.}

\textcolor{black}{Despite specific segmentation difficulties in shoulder muscles related to complex shapes and reduced sizes, our contributions show good performance with, in particular, excellent specificity (Tab.\ref{tab::sec4-tab-1}). In shoulder muscles, better segmentation results are highlighted for mid muscle regions (Fig.\ref{fig::sec4-2-fig-1}) where muscles appear bigger and well differentiated from surrounding tissues. Thus, we can assume that our approach could have very good performance for larger muscles with stable shapes like most of arm, forearm, thigh and leg muscles. Additionally, it provides interesting perspectives for other muscular disorders, for which objective and non-invasive biomarkers are required to effectively monitor both disease progression and treatment response.}

\textcolor{black}{At a research level, it could document effects of innovative treatments like genetic therapies for neuromuscular disorders \cite{ropars2020muscle} or improve the understanding of particular symptoms or diseases \cite{engstrom2007quadratus}. It could also be integrated into bio-mechanical models \cite{holzbaur2005model,blemker2005three} to help clinicians for intervention planning.}

\section{Conclusion}
\label{sec:sec5}

In this work, we successfully addressed automatic pathological shoulder muscle MRI segmentation for patients with obstetrical brachial plexus palsy by means of deep convolutional encoder-decoders. In particular, we studied healthy to pathological learning transferability by comparing different learning schemes in terms of model generalizability against large muscle shape, size, location, texture and injury variability. Moreover, convolutional encoder-decoder networks were expanded using VGG-16 encoders pre-trained on ImageNet to improve the accuracy reached by standard U-Net architectures. Our contributions were evaluated on four different shoulder muscles\textcolor{black}{: deltoid, infraspinatus, supraspinatus and subscapularis}. First, results clearly show that features extracted from unimpaired limbs are suited enough for pathological anatomies while acting as an efficient data augmentation strategy. Compared to transfer learning, combining healthy and pathological data for training provides the best segmentation accuracy together with outstanding delineation performance for muscle boundaries including insertion areas. Second, experiments reveal that convolutional encoder-decoders involving a pre-trained VGG-16 encoder strongly outperforms U-Net. Despite the non-medical nature of pre-training data, such deeper networks are able to efficiently discriminate individual muscles from surrounding anatomical structures. These conclusions offer new perspectives for the management of musculo-skeletal \textcolor{black}{disorders}, even if a small and heterogeneous \textcolor{black}{dataset} is available. The proposed approach can be easily extended to other muscle types and imaging modalities to provide \textcolor{black}{decision} support in various applications including neuro-muscular diseases, sports related injuries or any other muscle disorders. \textcolor{black}{Methodological perspectives on domain adaptation should deserve further investigation to take advantage of multi-centric data. Clinically, our method can be useful to distinguish between pathologies, evaluate the effect of treatments and facilitate surveillance of neuro-muscular disease course. It could be exploited together with bio-mechanical models to improve the understanding of complex pathologies and help clinicians to plan surgical interventions.}

\section*{Conflicts of interest}

\noindent None of the authors of this manuscript have any financial or personal relationships with other people or organizations that could inappropriately influence and bias this work.


\bibliography{conze-CMIG-2020-arXiv}

\end{document}